\pdfoutput=1

\PassOptionsToPackage{prologue,dvipsnames}{xcolor} 

\documentclass[11pt]{article}

\usepackage[final]{acl}
\usepackage{hyperref}
\usepackage{times}
\usepackage{latexsym}

\usepackage[T1]{fontenc}
\usepackage{CJKutf8}

\usepackage[utf8]{inputenc}

\usepackage{microtype}

\usepackage{inconsolata}

\usepackage{multirow}

\usepackage{listings}
\usepackage{tabularx}
\usepackage{ltablex}
\usepackage{longtable}
\usepackage{graphicx}
\usepackage{colortbl}
\usepackage{fontawesome5}

\usepackage{supertabular}
\usepackage{xltabular}
\usepackage{amsmath}
\usepackage{amssymb}
\usepackage{xspace}
\usepackage{mathrsfs}
\usepackage{bm,dutchcal}
\usepackage{float}
\usepackage{booktabs} 
\usepackage{makecell} 
\usepackage{amsmath} 

\usepackage{tcolorbox}
\tcbuselibrary{breakable, skins, listings}

\usepackage{marvosym}

\usepackage{adjustbox}
\usepackage{lineno}
\usepackage{array}
\usepackage{booktabs} 
\usepackage{makecell}
\usepackage[normalem]{ulem}

\usepackage[acronym]{glossaries}

\newcommand{\glsplain}[1]{%
  \ifglsused{#1}%
    {\glsentryshort{#1}}%
    {\glsentrylong{#1} (\glsentryshort{#1})\glsunset{#1}}%
}

\newacronym{3DVG}{3DVG}{3D Visual Grounding}
\newacronym{LLM}{LLM}{Large Language Model}
\newacronym{VLM}{VLM}{Vision Language Model}
\newacronym{LGSP}{LGSP}{Language-Guided Spatial Pruning}
\newacronym{MCDR}{MCDR}{Multi-View-Conditioned Description Reformulation}
\newacronym{LLM-Grounder}{LLM-Grounder}{Fine-tuned LLM-based Grounder}
\newacronym{MLP}{MLP}{Multilayer Perceptron}

\definecolor{custom_light_blue}{rgb}{0.85, 0.95, 1}
\definecolor{custom_light_pink}{rgb}{1, 0.85, 0.85}
\definecolor{custom_light_purple}{rgb}{0.98, 0.91, 0.973}
\definecolor{custom_light_purple_2}{rgb}{0.98, 0.91, 0.953}
\definecolor{custom_darkgreen}{rgb}{0.0, 0.5, 0.0}

\definecolor{gtred}{RGB}{220,20,60} 
\definecolor{m2mblue}{RGB}{0,102,204}

\usepackage{worldflags}
\definecolor{softgrey}{gray}{0.85}

\definecolor{Prune_Color}{RGB}{180,114,200}   
\definecolor{Ground_Color}{RGB}{230,150,190} 
\definecolor{DescriptReform_Color}{RGB}{225, 30, 86}

\definecolor{bg}{RGB}{248,248,255}
\definecolor{frame}{RGB}{80,80,180}
\definecolor{titlebg}{RGB}{60,60,150}
\definecolor{string}{RGB}{196,26,22}
\definecolor{keyword}{RGB}{0,102,204}
\definecolor{comment}{RGB}{0,140,0}
\definecolor{number}{RGB}{160,32,240}

\lstdefinestyle{mypython}{
    basicstyle=\ttfamily\footnotesize,
    keywordstyle=\color{keyword}\bfseries,
    stringstyle=\color{string},
    commentstyle=\color{comment}\itshape,
    numberstyle=\tiny\color{gray},
    numbers=left,
    stepnumber=1,
    numbersep=8pt,
    showstringspaces=false,
    breaklines=true,
    frame=none,
    columns=fullflexible
}

\DeclareRobustCommand{\PruneGroundTitle}{%
  \begingroup\normalfont
  \raisebox{-0.2em}{%
    \includegraphics[height=2em]{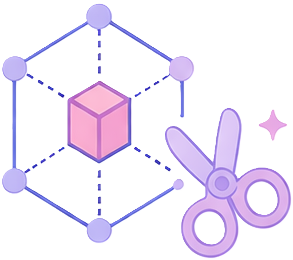}%
  }%
  \kern 0.0em 
  \textbf{\textcolor{Prune_Color}{Prune}}\textbf{\textcolor{Ground_Color}{Ground}}%
  \endgroup
}

\DeclareRobustCommand{\PruneGround}{%
  \begingroup\normalfont
  \raisebox{-0.2em}{%
    \includegraphics[height=1.2em]{images/Framework_Icon.PNG}%
  }%
  \kern 0.0em 
  \textbf{\textcolor{Prune_Color}{Prune}}\textbf{\textcolor{Ground_Color}{Ground}}%
  \endgroup
}

\title{\PruneGroundTitle: Plug-and-play Spatial \textcolor{Prune_Color}{Pruning} \\for 3D Visual \textcolor{Ground_Color}{Grounding}}

\author{Duc Cd$^{\text{\textcolor{Prune_Color}{\faStar}} \;1}$ \quad
Khai Le-Duc$^{\text{\textcolor{Prune_Color}{\faStar}}\;1,2,3}$ \quad Florent Draye$^{\;4}$ \quad Chris Ngo$^{\;1}$ 
\\ {\bf Terry Jingchen Zhang$^{\;3,6}$ \quad Bernhard Sch\"{o}lkopf$^{\;4,5}$ \quad Zhijing Jin$^{\;2,3,4}$}\\
$^1$ \resizebox{!}{0.7em}{\worldflag{SG}} Knovel Engineering Lab \enspace
$^2$ \resizebox{!}{0.7em}{\worldflag{CA}} University of Toronto \enspace
$^3$ \resizebox{!}{0.7em}{\worldflag{CA}} Vector Institute\\
$^4$ \resizebox{!}{0.7em}{\worldflag{DE}} MPI for Intelligent Systems, T{\"u}bingen 
$^5$ \resizebox{!}{0.7em}{\worldflag{DE}} ELLIS Institute Tübingen\enspace\\
$^{6}$ \resizebox{!}{0.7em}{\worldflag{GB}} University of Oxford
\\$^{\text{\textcolor{Prune_Color}{\faStar}}}$Co-first authors\\
\textcolor{Prune_Color}{\faEnvelope}  \texttt{duccd@hanyang.ac.kr} \quad
\textcolor{Prune_Color}{\faEnvelope} \texttt{duckhai.le@mail.utoronto.ca}\\
{\faGithub}   \href{https://github.com/leduckhai/PruneGround}{leduckhai/\textbf{\textcolor{Prune_Color}{Prune}}\textbf{\textcolor{Ground_Color}{Ground}}} 
\quad {\textcolor{Prune_Color}{\faGlobe}} \href{https://leduckhai.github.io/PruneGround/}{Project Page} 
}

\begin{document}
\maketitle
\begin{abstract}
3D Visual Grounding (3DVG) aims to localize target objects in 3D scenes given natural language descriptions. Existing approaches typically perform reasoning over the entire scene, leading to ambiguous predictions and high computational cost, especially in cluttered environments. We observe that many referential expressions rely on local spatial context and often correspond to restricted spatial regions rather than the full scene. Motivated by this insight, we propose \PruneGround, an effective plug-and-play framework for 3DVG built upon three key components. First, we introduce \textit{Language-Guided Spatial Pruning} (LGSP), which leverages a frozen Vision Language Model (VLM) to identify language-relevant regions, thereby reducing spatial computation and grounding candidates in the narrower search space. Second, we propose \textit{MultiView-Conditioned Description Reformulation} (MCDR), which decomposes complex expressions into simplified target--anchor relations and augments missing spatial cues through multi-view reasoning. Finally, we propose \textit{LLM-Grounder}, which repurposes a detection-pretrained spatial LLM into a language-conditioned grounding model by aligning point cloud and linguistic representations within the pruned region. Extensive experiments on the three most popular point cloud benchmarks demonstrate that our method achieves \textbf{state-of-the-art} results on all three ScanRefer settings  and on 9 out of 10 Nr3D/Sr3D settings. 
\end{abstract}

\thispagestyle{plain}
\pagestyle{plain}

\addtocontents{toc}{\protect\setcounter{tocdepth}{-1}}

\section{Introduction}
\label{sec:intro}
\glsplain{3DVG} focuses on linking natural language expressions to their corresponding objects in complex 3D environments, demanding robust language understanding and spatial reasoning, and serving as a key component in applications such as augmented reality~\cite{chen2020scanrefer, Ma2023AnEO}, embodied navigation~\cite{Zhou2023NavGPTER, Chen2022ThinkGA, Huang2022ASSISTERAN}, intelligent agents~\cite{Calisto2023AssertivenessbasedAC, Zhang2023AppAgentMA}, robotic perception~\cite{Chen2023CLIP2SceneTL, Kong2023RethinkingRV, Kong2023Robo3DTR, Lai2023XVOGV}, and autonomous driving~\cite{Cui2023DriveAY, Feng2023DenseRL}.
In this task, most existing methods adopt a \textit{segment-then-select} paradigm~\cite{Yuan2021InstanceReferCH, Mane2025Ges3ViGIP, zhang2024dlisa, Xu2024MultiAttributeIM, Chang2024MiKASAM}, where 3D object proposals are first extracted via scene-level segmentation and then matched to the textual query. Despite strong benchmark performance, such \textit{two-stage} approaches suffer from inherent limitations, including \textit{error propagation} from imperfect segmentation and high computational cost, often redundant in cluttered scenes with fine-grained relational cues.

\begin{figure}[t]
    \centering
    \includegraphics[width=\linewidth]{images/introduction_1.pdf}
    \caption{\textbf{Intuition of our proposed method, inspired by human spatial cognition} (see Appendix Sec. \ref{sec:cognitive_motivation}). Given a language description, the grounding model, as humans do, should infer the target object by attending \textit{only} to spatially relevant regions guided by anchor objects and spatial relations, rather than reasoning over \textit{all} candidate objects in the scene (e.g., all chairs).}
    \label{fig:introduction_1}
    ~\vspace{-0.5cm}
\end{figure}

Given a language description for localizing a target object, the contextual cues typically correspond to only a restricted region of the 3D scene. As illustrated in Fig.~\ref{fig:introduction_1}, instead of reasoning over all candidate objects (e.g., all chairs in the scene), the grounding model should focus \textit{only} on spatially relevant areas guided by the anchor objects and spatial relations described in the utterance. This mirrors human spatial cognition, where people interpret referring expressions by using salient anchor objects and relational cues to narrow attention to the most relevant region of a scene rather than exhaustively comparing every possible object.

Additionally, human utterances are often free-form and exhibit ambiguity, incompleteness, and implicit spatial reasoning. Spatial relations may be partially specified or omitted, relying on assumptions that are intuitive for humans but challenging for models. This variability complicates grounding, as models must resolve ambiguity and infer missing geometric cues.

Consider the red chair in Fig. \ref{fig:introduction_1}. The utterance may rely on the door to identify the target, yet in point clouds the door can be ambiguous, e.g. sharing color with the wall or partially missing due to occlusion and capture artifacts. Models that depend on the door as an anchor are thus prone to confusion. To improve localization, additional contextual cues are needed. Relations with nearby objects, such as a desk or bookshelf, provide complementary constraints that help disambiguate candidates and yield more stable bounding box estimation in cluttered or imperfect scenes.

Motivated by these problems, we propose \PruneGround, a novel plug-and-play,  \textit{\textbf{\textcolor{Prune_Color}{prune}}-then-\textbf{\textcolor{Ground_Color}{ground}}} framework that leverages a \glsplain{VLM} to prune irrelevant regions of a 3D scene, reformulate descriptions, and optimize a \glsplain{LLM} for effective grounding. By focusing computation on language-relevant regions, our framework enables a target-centric and more accurate grounding process.


We devise a \textit{three-stage} pipeline for effective \glsplain{3DVG}:

(i) \textbf{\textcolor{Prune_Color}{\glsplain{LGSP}}}: We first render global multi-view RGB and depth images of the 3D scene and leverage a \glsplain{VLM} to identify regions that are semantically relevant to the input query. This step effectively narrows the spatial search space and reduces unnecessary computation over irrelevant areas (Sec.~\ref{LGSP}).

(ii) \textbf{\textcolor{DescriptReform_Color}{\glsplain{MCDR}}}: We further render side views and cropped top-view regions centered on the pruned areas, and employ the \glsplain{VLM} to decompose complex multi-anchor expressions into structured single-anchor statements. In addition, we augment the language descriptions with explicit spatial relations between the target and surrounding anchors that may be implicit or absent in the original utterance (Sec.~\ref{MCDR}).

(iii) \textbf{\textcolor{Ground_Color}{\glsplain{LLM-Grounder}}}: Finally, we repurpose a detection-pretrained spatial \glsplain{LLM} for language-conditioned grounding by aligning point cloud and linguistic representations within the pruned region and adapting the decoding scheme to produce the target prediction in a structured language form (Sec.~\ref{finetuneLLM}).

By simply integrating \glsplain{VLM}-based scene pruning and description reformulation with an effective \glsplain{LLM}-based grounding framework, our plug-and-play \PruneGround\, achieves \textbf{state-of-the-art} performance across the three most popular \glsplain{3DVG} benchmarks - ScanRefer~\cite{chen2020scanrefer}, Nr3D, and Sr3D from ReferIt3D~\cite{Achlioptas2020ReferIt3DNL}. Notably, it demonstrates strong generalization by delivering state-of-the-art results on all three ScanRefer settings (Tab.~\ref{tab:table_1}) and on 9 out of 10 Nr3D/Sr3D settings (Tab.~\ref{tab:table_2}).

\section{Related Works}

\paragraph{Two-stage vs. One-stage \glsplain{3DVG}.} Pioneering ScanNet-based benchmarks such as ScanRefer~\cite{chen2020scanrefer} and ReferIt3D~\cite{Achlioptas2020ReferIt3DNL,Dai2017ScanNetR3} have advanced \glsplain{3DVG}. Early supervised methods typically use separate language and scene encoders followed by cross-modal fusion.

(i) \textit{Two-stage} approaches~\cite{Zhao20213DVGTransformerRM, Achlioptas2020ReferIt3DNL, Bakr2022LookAA, Chen2022LanguageCS, Yang2021SAT2S, Yuan2021InstanceReferCH, Zhu20233DVisTAPT, Mane2025Ges3ViGIP, Wang2025AugReferA3, Zhang2024TowardsCL, Chang2024MiKASAM, Yang2023ExploitingCO, Bakr2023CoT3DRefCD, zhang2024dlisa} follow a \textit{proposal-and-selection} paradigm: a 3D detector or segmentor first generates candidate proposals, then the proposal best matching the query is selected. By decoupling object localization and language grounding, they leverage strong geometric priors and mature detection pipelines through explicit candidate reasoning \cite{le2026vision}.

(ii) In contrast, \textit{one-stage} approaches~\cite{Jain2021BottomUT, jain2025unifying, Luo20223DSPSS3, Qian2024MultibranchCL, Wang2024G3LQMH, Wu2022EDAET, Guo2025TextguidedSV, Yuan2023VisualPF, Wu20233DSTMNDS} directly localize the target in the 3D scene conditioned on the query. This enables \textit{end-to-end} optimization and tighter language-geometry interaction, while avoiding error propagation from explicit proposal generation.

\paragraph{\glsplain{LLM}- and \glsplain{VLM}-based \glsplain{3DVG}.} To enhance language understanding in \glsplain{3DVG}, recent works incorporate \glsplain{LLM}s~\cite{Guo2023ViewReferGT, Yin2023LAMMLM, Zhan2023Mono3DVG3V, Yuan2023VisualPF, Feng2024NaturallyS3, Zhu2024ScanReasonE3, Mi2025LanguagetoSpacePF, Liu2025ReasonGrounderLH} and \glsplain{VLM}s~\cite{Xu2024VLMGrounderAV, Wang2025LIBALI, Cao2025FromTT, Li2025ZeroShot3V, Zhu2025Grounding3O, Jung2025PanoGrounderB2, Jin2025SPAZERSP} into the grounding pipelines.

(i) \textit{Zero-shot} approaches exploit pretrained \glsplain{VLM}s for open-vocabulary \glsplain{3DVG} without task-specific 3D supervision. ZSVG3D~\cite{Yuan2023VisualPF} uses \glsplain{LLM}-driven visual programming and modular 3D reasoning; VLM-Grounder~\cite{Xu2024VLMGrounderAV} performs iterative grounding over stitched multi-view images and projects ensembles to 3D boxes; and SeeGround~\cite{Li2024SeeGroundSA} converts 3D scenes into query-aligned rendered views with spatial textual descriptions to match \glsplain{VLM} inputs.

\textit{(ii) Fine-tuned} approaches embed \glsplain{VLM}s/\glsplain{LLM}s into 3D grounding networks with task-specific box decoders. LIBA~\cite{Wang2025LIBALI} mitigates point-language granularity mismatch via multi-granularity alignment, dynamic cross-modal adapters, and \glsplain{LLM}-guided hierarchical reasoning. ReGround3D~\cite{Zhu2024ScanReasonE3} introduces a reasoning-oriented 3D grounding benchmark and a \glsplain{VLM}-based chain-of-grounding framework that interleaves reasoning and localization.


\paragraph{Structured language reasoning for \glsplain{3DVG}.} Accurate \glsplain{3DVG} requires reasoning over attributes, relations, and viewpoint-dependent expressions. Recent works improve language grounding by converting free-form utterances into structured semantic forms \cite{le2025s}. ViewRefer~\cite{Guo2023ViewReferGT} uses \glsplain{LLM}s to enrich queries with view-aware templates, while EDA~\cite{Wu2022EDAET} decomposes descriptions into fine-grained semantic elements for dense alignment with point cloud instances. ViewSRD~\cite{Huang_2025_ICCV} restructures multi-anchor expressions into simpler single-anchor statements to disentangle target-anchor relations.

Similarly, we decompose multi-anchor utterances into single-anchor relations. Unlike purely text-based restructuring~\cite{Guo2023ViewReferGT}, our method renders multi-view images and uses a \glsplain{VLM} to decouple relations, remove redundancies, and supplement missing geometric cues, yielding richer perspective-aware descriptions.

\section{\PruneGround}
\begin{figure*}[t]
    \centering
    \includegraphics[width=\linewidth]{images/main_pipeline.pdf}
    \caption{
\textbf{Overview of our} \PruneGround\ pipeline, consisting of three \textit{plug-and-play} components: \textbf{\textcircled{\scriptsize i}} \textbf{\textcolor{Prune_Color}{\glsplain{LGSP}}} to remove irrelevant regions (Sec.~\ref{LGSP}), \textbf{\textcircled{\scriptsize ii}} \textbf{\textcolor{DescriptReform_Color}{\glsplain{MCDR}}} to simplify the description while adding relational spatial information (Sec.~\ref{MCDR}), and \textbf{\textcircled{\scriptsize iii}} \textbf{\textcolor{Ground_Color}{\glsplain{LLM-Grounder}}} that repurposes a detection-pretrained spatial \glsplain{LLM} for language-conditioned grounding on the pruned scene and reformulated description (Sec.~\ref{finetuneLLM}).}
    \label{fig:main_pipeline}
    ~\vspace{-0.5cm}
\end{figure*}
\label{sec:pruneground}
\subsection{Overall Architecture}

As shown in Fig.~\ref{fig:main_pipeline}, given a 3D scene represented by point coordinates and color features $\mathcal{S} \in \mathbb{R}^{N \times 6}$ and a \textbf{\textcolor{orange}{natural language query $\mathcal{Q}$}}\footnote{The colors of the \textit{variables} and \textit{components} in Sec. \ref{sec:pruneground} correspond to their colors in Fig.~\ref{fig:main_pipeline}}, our goal is to localize the referred target object. Instead of grounding over the entire scene, we propose a \textit{three-stage} framework that progressively reduces spatial redundancy and refines language understanding.

(i) We introduce \textbf{\textcolor{Prune_Color}{\glsplain{LGSP}}} (Sec.~\ref{LGSP}), where multi-view renderings (top and oblique views) are aligned with the query using a pretrained \glsplain{VLM} to identify semantically relevant regions. Irrelevant 3D points are removed, yielding a pruned scene that reduces clutter and narrows the search space.

(ii) We present \textbf{\textcolor{DescriptReform_Color}{\glsplain{MCDR}}} (Sec.~\ref{MCDR}). Using top and side views of the pruned scene, we identify target and anchor objects and reformulate complex expressions into structured \textit{single-target}$\rightarrow$\textit{single-anchor} relations. Redundant statements are filtered, and the query is enriched with geometric cues inferred from multi-view semantics.

(iii) Given the pruned scene and reformulated description, we develop \textbf{\textcolor{Ground_Color}{\glsplain{LLM-Grounder}}} (Sec.~\ref{finetuneLLM}), a language-conditioned grounding model repurposed from a detection-pretrained spatial \glsplain{LLM}, which jointly reasons over language and point cloud representations to predict the target object.

\subsection{Language-Guided Spatial Pruning (\textcolor{Prune_Color}{LGSP})}
\label{LGSP}

Let an indoor 3D scene represented as a point cloud with coordinates and color, $\mathcal{S} \in \mathbb{R}^{N \times 6}$, where the $x$-$y$ plane aligns with the floor and the $z$-axis points upward, opposite gravity. This canonical alignment enables consistent multi-view rendering and spatial reasoning.

\paragraph{~\textbf{Global Multi-View Rendering.}}
The \textbf{\textcolor{Prune_Color}{\glsplain{LGSP}}} module first renders a set of structured multi-view observations from $\mathcal{S}$. Specifically, we generate one top-view image that serves as a canonical spatial layout of the scene, together with four oblique views. The top view captures the global spatial arrangement along the horizontal plane but may lack semantic details of vertically structured objects. To complement this, each oblique view provides an RGB image encoding object semantics and a corresponding depth map encoding geometric structure.

To construct the oblique views, we compute the axis-aligned bounding box (AABB) of the scene to define its spatial extent. The camera centers are placed at the midpoints of the four upper edges of the AABB. Each camera is oriented toward the center of the opposite vertical face, ensuring that the field of view covers the corresponding side of the scene (see Appendix Fig. \ref{fig:sup_global_cam}). RGB and depth images are rendered simultaneously under this configuration. Implementation details of the camera intrinsics and rendering settings are provided in the Appendix Sec. \ref{sec:camera_setup}. The final multi-view representation thus consists of one top-view image, four oblique RGB images, and four corresponding oblique depth maps, as shown in Fig.~\ref{fig:main_pipeline}.

\paragraph{~\textbf{Language-Guided Region Selection.}}
The rendered multi-view images, together with the \textbf{\textcolor{orange}{language query $\mathcal{Q}$}}, are fed into a \textbf{\textcolor{cyan}{frozen \glsplain{VLM}}}. Conditioned on the textual query, the \glsplain{VLM} analyzes the semantic and spatial correspondences across views and predicts a relevant \textbf{\textcolor{green}{2D bounding region}} on the top-view image that is most consistent with the description.
~\vspace{-0.15cm}
\begin{equation}
\begin{aligned}
\textbf{\textcolor{green}{$\mathbf{bbox}^{2D}$}} 
&= (x_{\min}, x_{\max}, y_{\min}, y_{\max}) \\
&= \textbf{\textcolor{cyan}{$\mathrm{VLM}$}} (\mathcal{V}_{\text{global}}, \textbf{\textcolor{orange}{$\mathcal{Q}$}})
\end{aligned}
~\vspace{-0.15cm}
\end{equation}
where $\mathcal{V}_{\text{global}}$ denotes the \textit{global multi-view} input, and $(x_{\min}, x_{\max}, y_{\min}, y_{\max})$ define the boundaries of the retained region. 

Details of the structured prompting for the \glsplain{VLM} are provided in Fig. \ref{fig:prompt_2d_bbox_estimation} of Appendix Sec. \ref{sec:VLM_query_prompt}.

\paragraph{~\textbf{Post-processing and 3D Projection.}}
To ensure spatial stability, we apply post-processing to the predicted \textbf{\textcolor{green}{2D bounding box}} in the top view by enforcing a \setlength{\fboxsep}{2pt}
\textit{minimum area constraint}. If the box falls below this threshold, it is proportionally expanded to maintain sufficient coverage. The refined box is then projected back into 3D space. Since the top view lies on the $x$-$y$ plane (with a default resolution of $512 \times 512$), we retain the full extent along the $z$-axis and select points whose $(x, y)$ coordinates lie within the box, discarding the rest to obtain the pruned point cloud $\mathcal{S}'$.

For compact scenes where the area or pruned region is smaller than $1\,\text{m} \times 1\,\text{m}$, we retain the entire scene or expand the region accordingly to avoid unnecessary information loss.

\subsection{Multi-View-Conditioned Description Reformulation (\textbf{\textcolor{DescriptReform_Color}{MCDR}})}
\label{MCDR}

Natural language descriptions in \glsplain{3DVG} are inherently human-centric and often reflect intuitive reasoning rather than complete spatial specifications. As a result, relevant objects or spatial relations may be omitted or only implicitly expressed. While ViewSRD~\cite{Huang_2025_ICCV} decomposes expressions into target-anchor pairs to facilitate relational learning, it is purely text-driven and does not condition on visual semantics. Consequently, it may miss spatial cues that are visually evident but linguistically absent. To address this, we propose an \textbf{\textcolor{DescriptReform_Color}{\glsplain{MCDR}}} module that restructures descriptions into target-anchor relations under multiview visual conditioning. By incorporating semantic cues from top and side views of the pruned region, our method simplifies queries into \textit{single-target}$\rightarrow$\textit{single-anchor} statements while recovering missing spatial information from scene geometry.

\paragraph{\textbf{Local Multi-View Rendering.}}
For the pruned scene $\mathcal{S}'$ represented by a \textit{3D bounding box}, we render two side-view RGB images from the $x$- and $y$-axis directions of the \textit{selected top-view image's bounding box}. For the camera aligned with the $x$-axis, the viewing direction is set to pass through the centers of the two faces parallel to the $y$-axis. The camera is then gradually elevated to reduce occlusions from surrounding furniture. To further mitigate occlusions from structural elements (e.g., walls), the camera orientation is constrained to face outward from the center of the original scene. The camera along the $y$-axis is configured using the same procedure. In this step, depth maps are discarded to simplify the input and emphasize semantic content. A visualization of the camera configuration and rendering pipeline is provided in Fig. \ref{fig:sup_local_cam} of Appendix Sec. \ref{sec:camera_setup}.

\paragraph{\textbf{Description Simplification and Augmentation.}}
Given rendered local multi-view images $\mathcal{V}_{\text{local}}$ and the \textbf{\textcolor{orange}{original description $\mathcal{Q}$}}, we continue to leverage a \textbf{\textcolor{cyan}{frozen \glsplain{VLM}}} to perform \textit{structured} query reformulation:
\begin{equation}
(t, \mathcal{T}, \mathcal{A}, \hat{\mathcal{S}})
=
\textbf{\textcolor{cyan}{$\mathrm{VLM}$}}(\mathcal{V}_{\text{local}}, \textbf{\textcolor{orange}{$\mathcal{Q}$}}).
\end{equation}

The \textbf{\textcolor{cyan}{\glsplain{VLM}}} decomposes the \textbf{\textcolor{orange}{query $\mathcal{Q}$}} into four components:
(i) the target object name $t$,
(ii) a set of attributes of the target object $\mathcal{T}$, which describe properties such as color, material, and shape,
(iii) a set of anchor object names $\mathcal{A}$, and
(iv) a set of target--anchor relational statements $\hat{\mathcal{S}}$,
each expressed as:
$\langle \text{target}, \text{target attributes} \rangle + \langle \text{spatial relation} \rangle + \langle \text{anchor}, \text{anchor attributes} \rangle.$

By prompting the \textbf{\textcolor{cyan}{\glsplain{VLM}}} to augment relational information using the local multi-view images, the resulting relational statements incorporate not only reformulated content from the \textbf{\textcolor{orange}{original description}} but also additional target--anchor pairs. Specifically, anchors that do not explicitly appear in the original description but are observable in the local multi-view images may be introduced, leading to enriched relational supervision grounded in visual context. Details of the structured prompting strategy for the VLM query are provided in Fig. \ref{fig:prompt_MCDR} of Appendix Sec. \ref{sec:VLM_query_prompt}.

\subsection{LLM-based Grounder (\textbf{\textcolor{Ground_Color}{LLM-Grounder}})}
\label{finetuneLLM}
Given a reformulated referring expression and a pruned 3D point cloud as input, we build upon the SpatialLM framework~\cite{SpatialLM}, originally developed for unconditional 3D object detection, and repurpose it for language-conditioned grounding. The architecture follows an \textit{\textcolor{cyan}{Encoder}--\textcolor{cyan}{Projector}--\textcolor{red}{LLM}} design.

A point cloud encoder (Sonata \cite{wu2025sonata}, a PTv3 variant with 108.8M parameters \cite{wu2024point}) first extracts geometric features from the pruned scene. Unlike the original SpatialLM, which processes point clouds without linguistic conditioning, we inject the referring expression into the multimodal sequence to enable target-aware reasoning. Specifically, the geometric features are projected via a Multilayer Perceptron (MLP) into the language embedding space and concatenated with the text token embeddings of the referring expression, forming a unified multimodal input sequence. The \glsplain{LLM} (Qwen2.5-0.5B \cite{qwen2}) processes this sequence and predicts a single target object matching the referring expression, in a structured language format:
\begin{equation}
\texttt{<}c,\; p_x, p_y, p_z,\; s_x, s_y, s_z\texttt{>}
\end{equation}
where $c$ denotes the object name, $(p_x, p_y, p_z)$ the center position, and $(s_x, s_y, s_z)$ the size of the bounding box along each axis.
This contrasts with the enumerative output of the original SpatialLM, which produces structured tokens for all detected objects in the scene. Finally, the predicted bounding box parameters are mapped back to the original scene coordinate space to obtain the final grounding result.

\section{Experiments}
\subsection{Experimental Settings}

\paragraph{\textbf{Datasets.}} We evaluate our framework on three widely used \glsplain{3DVG} benchmarks. 
(i) \textbf{ScanRefer} contains 51,583 human-written referring expressions for 800 ScanNet scenes~\cite{Dai2017ScanNetR3}. Following the official split, we use 36,665 samples for training and 9,508 for validation. Based on object uniqueness within a scene, the dataset is divided into \textit{unique} (the target class appears once) and \textit{multiple} (the target class appears multiple times) subsets.
(ii) \textbf{Nr3D} consists of 37,842 human-written referring expressions grounded to annotated objects across 641 ScanNet scenes (511 for training and 130 for validation), covering 76 object classes. Each expression refers to a target object among same-class distractors. For evaluation, the dataset is partitioned into \textit{easy} (one distractor) and \textit{hard} (multiple distractors) subsets. Expressions are further categorized as \textit{view-dependent} or \textit{view-independent}, depending on whether grounding requires a specific viewpoint.
(iii) \textbf{Sr3D} contains 83,570 automatically generated referring expressions from 1,018 training and 255 validation ScanNet scenes. Unlike Nr3D, Sr3D uses predefined templates and relies purely on spatial relations to distinguish same-class objects. We follow the standard evaluation protocol consistent with Nr3D.

\paragraph{Implementation Details.}
We use \textit{Qwen2.5-VL-3B-Instruct}~\cite{Bai2025Qwen25VLTR} as the frozen \glsplain{VLM} for both \textbf{\textcolor{Prune_Color}{\glsplain{LGSP}}} and \textbf{\textcolor{DescriptReform_Color}{\glsplain{MCDR}}}. The top-view, oblique-view, and side-view images are rendered at a resolution of $512 \times 512$ pixels with a field of view (FoV) of $90^\circ$ (see Fig. \ref{fig:sup_global_cam} and Fig. \ref{fig:sup_local_cam} of Appendix Sec. \ref{sec:camera_setup}).  \textbf{\textcolor{Ground_Color}{\glsplain{LLM-Grounder}}} adopts the Qwen-based SpatialLM~\cite{SpatialLM} (Qwen2.5-0.5B) and initialize from its pre-trained checkpoint. Following \citet{SpatialLM}, we shift the point cloud and target bounding box coordinates to ensure non-negative values, then quantize them into 1,280 bins at 2.5cm resolution. The LLM predicts integer values, which are mapped back to continuous coordinates via inverse transformation. All experiments are implemented in PyTorch and run on an  NVIDIA A100 80~GB GPU. Further details of the pre-trained model initialization and training procedure are provided in Appendix Sec. \ref{sup: training}.

\subsection{Experimental Results}
\paragraph{\textbf{ScanRefer dataset}.}
Tab. ~\ref{tab:table_1} compares our method with prior approaches on the ScanRefer benchmark. On the challenging \emph{Multiple} subset, where multiple candidates share the same class as the target, our pruning strategy effectively removes objects in irrelevant regions, achieving 57.6\% Acc\texttt{@}0.25 and 49.4\% Acc\texttt{@}0.5. As a result, our method achieves 63.8\% Acc\texttt{@}0.25 and 56.1\% Acc\texttt{@}0.5 overall, establishing a new \textbf{state of the art in all three settings}: \textit{Unique}, \textit{Multiple}, \textit{Overall}.
 
\begin{table*}[t]
\centering
\small
\setlength{\tabcolsep}{2pt}
\begin{tabular}{l l | cc | cc | cc}
\toprule
\multirow{2}{*}{\textbf{Method}} & \multirow{2}{*}{\textbf{Venue}} 
& \multicolumn{2}{c|}{\textbf{Unique}} 
& \multicolumn{2}{c|}{\textbf{Multiple}} 
& \multicolumn{2}{c}{\textbf{Overall}} \\
\cmidrule(lr){3-4} \cmidrule(lr){5-6} \cmidrule(lr){7-8}
 &  & Acc\texttt{@}0.25 & Acc\texttt{@}0.5 & Acc\texttt{@}0.25 & Acc\texttt{@}0.5 & Acc\texttt{@}0.25 & Acc@0.5 \\
\midrule
ScanRefer~\cite{chen2020scanrefer} & ECCV & 67.6 & 46.2 & 32.1 & 21.3 & 40.0 & 26.1 \\
MVT~\cite{Huang2022MultiViewTF} & CVPR & 77.7 & 66.5 & 31.9 & 25.3 & 40.8 & 33.3 \\
3D-SPS~\cite{Luo20223DSPSS3} & CVPR & 84.1 & 66.7 & 40.3 & 29.8 & 48.8 & 37.0 \\
EDA~\cite{Wu2022EDAET} & CVPR & 85.8 & 68.6 & 49.1 & 37.6 & 54.6 & 42.3 \\
ConcreteNet~\cite{Unal2023FourWT} & ECCV & 86.4 & 82.1 & 42.4 & 38.4 & 50.6 & 46.5 \\
Chat-Scene~\cite{Huang2023ChatSceneB3} & NeurIPS & 89.6 & \underline{82.5} & 47.8 & 42.9 & 55.5 & 50.2 \\
VPP-Net~\cite{Shi2024ViewpointAwareVG} & CVPR & 86.1 & 67.1 & 50.3 & 39.0 & 55.7 & 43.3 \\
$G^3$-LQ~\cite{Wang2024G3LQMH} & CVPR & 88.6 & 73.3 & 50.2 & 39.7 & 56.0 & 44.7 \\
$MA^2$TransVG~\cite{Xu2024MultiAttributeIM} & CVPR & 86.3 & 74.1 & 53.8 & 41.4 & 57.9 & 45.7 \\
MCLN~\cite{Qian2024MultibranchCL} & ECCV & 86.9 & 72.7 & 52.0 & 40.8 & 57.2 & 45.5 \\
PQ3D~\cite{Zhu2024Unifying3V} & ECCV & 86.7 & 78.3 & 51.5 & 46.2 & 57.0 & 51.2 \\
VGMamba~\cite{zhu2025vgmamba} & ICCV & \textbf{91.9} & 79.6 & 54.8 & \underline{47.6} & 60.0 & \underline{53.9} \\
PanoGrounder~\cite{Jung2025PanoGrounderB2} & ECCV & 84.3 & -- & \underline{55.3} & -- & \underline{61.0} & -- \\
\rowcolor{custom_light_purple_2}
\PruneGround \, \textbf{(Ours)} &  
& \underline{90.2} & \textbf{84.7} 
& \textbf{57.6} & \textbf{49.4} 
& \textbf{63.8} & \textbf{56.1} \\
\bottomrule
\end{tabular}
\caption{
Grounding accuracy (\%) on \textbf{ScanRefer} (Acc\texttt{@}0.25 / 0.5, IoU).
\textbf{Bold}: best, \underline{underline}: second-best.
}
\label{tab:table_1}
\end{table*}
\begin{table*}[t]
\centering
\small
\setlength{\tabcolsep}{2pt}
\begin{tabular}{l l | ccccc | ccccc}
\toprule
\multirow{2}{*}{\textbf{Method}} & \multirow{2}{*}{\textbf{Venue}} 
& \multicolumn{5}{c|}{\textbf{Nr3D}} 
& \multicolumn{5}{c}{\textbf{Sr3D}} \\
\cmidrule(lr){3-7} \cmidrule(lr){8-12}
 &  & Easy & Hard & View-D & View-I & Overall 
 & Easy & Hard & View-D & View-I & Overall \\
\midrule

ReferIt3D~\cite{Achlioptas2020ReferIt3DNL} & ECCV 
& 43.6 & 27.9 & 32.5 & 37.1 & 35.6 
& 44.7 & 31.5 & 39.2 & 40.8 & 40.8 \\

ViL3DRel~\cite{Chen2022LanguageCS} & NeurIPS 
& 70.2 & 57.4 & 62.0 & 64.5 & 64.4 
& 74.9 & 67.9 & 63.8 & 73.2 & 72.8 \\

3D-VisTA~\cite{Zhu20233DVisTAPT} & ICCV 
& 65.9 & 49.4 & 53.7 & 59.4 & 57.5 
& 72.1 & 63.6 & 57.9 & 70.1 & 69.6 \\

MIKASA~\cite{Chang2024MiKASAM} & CVPR 
& 69.7 & 59.4 & 65.4 & 64.0 & 64.4 
& 78.6 & 67.3 & \textbf{70.4} & 75.4 & 75.2 \\

$G^3$-LQ~\cite{Wang2024G3LQMH} & CVPR 
& -- & 50.7 & -- & -- & 58.4 
& -- & 66.3 & -- & -- & 73.1 \\

$MA^2$TransVG~\cite{Xu2024MultiAttributeIM} & CVPR 
& -- & 57.6 & -- & -- & 65.2 
& -- & 69.3 & -- & -- & 73.9 \\

PQ3D~\cite{Zhu2024Unifying3V} & ICCV 
& 73.3 & 56.7 & 60.7 & 67.0 & 64.9 
& 78.8 & 68.2 & 51.5 & 76.7 & 75.6 \\

LIBA~\cite{Wang2025LIBALI} & AAAI 
& -- & 57.2 & 60.3 & -- & 64.5 
& -- & 70.2 & 61.7 & -- & 75.8 \\

ViewSRD~\cite{Huang_2025_ICCV} & ICCV 
& 75.3 & 64.8 & 68.6 & 70.6 & 69.9 
& 78.3 & 70.6 & \underline{69.0} & 76.2 & 76.0 \\

VGMamba~\cite{zhu2025vgmamba} & ICCV 
& -- & 61.4 & -- & -- & 68.3 
& -- & \underline{74.4} & -- & -- & \underline{81.3} \\

PanoGrounder~\cite{Jung2025PanoGrounderB2} & ECCV 
& \underline{82.2} & \underline{67.2} & \underline{70.5} & \underline{76.3} & \underline{74.6} 
& \underline{81.3} & 74.2 & 60.5 & \underline{80.0} & 79.1 \\

\rowcolor{custom_light_purple_2}
\PruneGround \, \textbf{(Ours)} &  
& \textbf{84.6} & \textbf{68.8} & \textbf{71.2} & \textbf{76.6} & \textbf{75.1} 
& \textbf{83.7} & \textbf{76.5} & 61.7 & \textbf{82.4} & \textbf{81.5} \\

\bottomrule
\end{tabular}
\caption{
Grounding accuracy (\%) on \textbf{Nr3D} and \textbf{Sr3D} (Acc\texttt{@}0.5, IoU). \textbf{Bold}: best, \underline{underline}: second-best}

\label{tab:table_2}
\end{table*}

\begin{figure*}[t]
     \centering
     \includegraphics[width=0.95\linewidth]{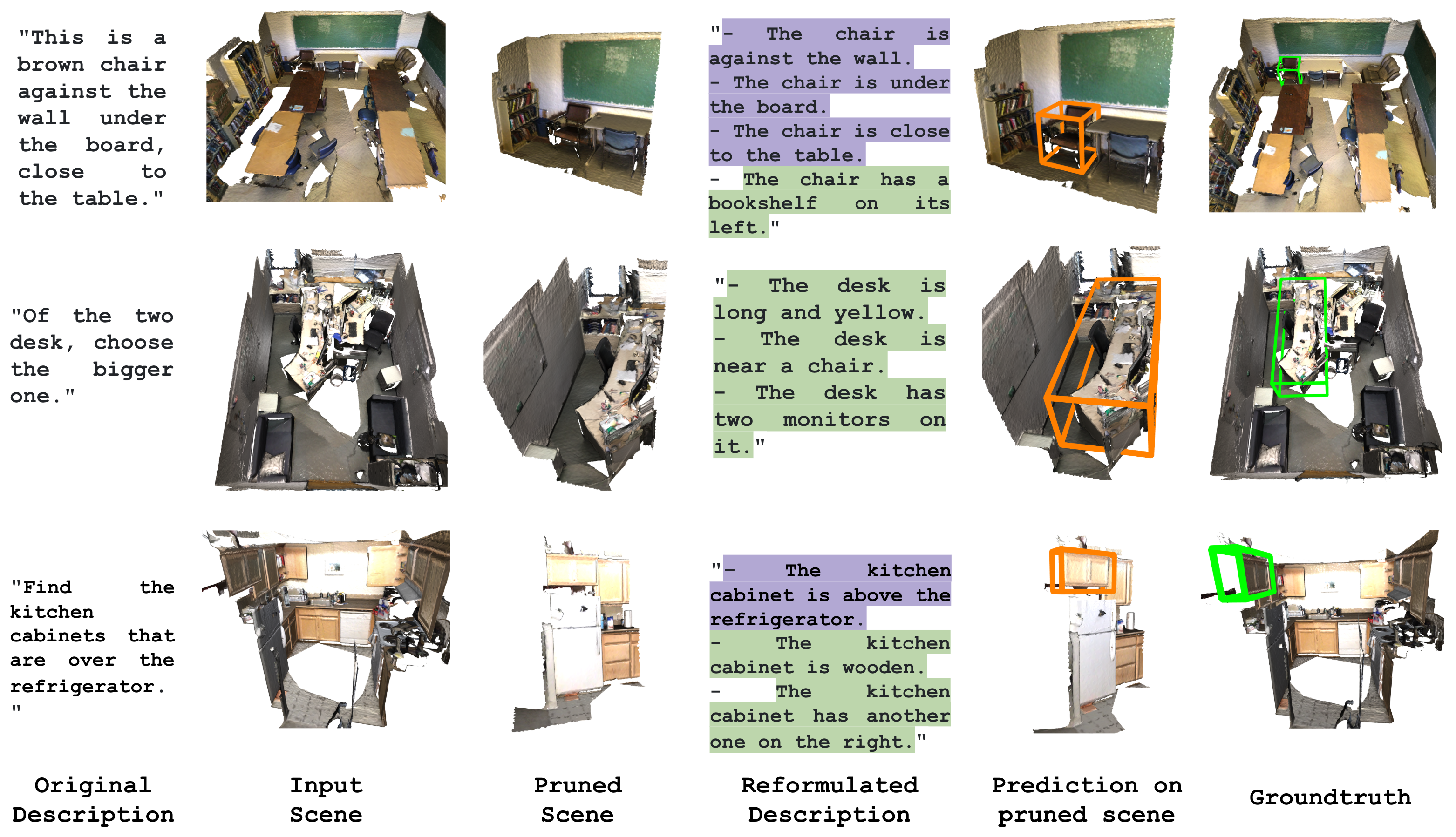}
     \caption{
\textbf{Qualitative results on ScanRefer, Nr3D, and Sr3D.} The first column shows original utterances with varying complexity. The third column presents scenes pruned by \textbf{\textcolor{Prune_Color}{\glsplain{LGSP}}}. The fourth column shows reformulated descriptions from \textbf{\textcolor{DescriptReform_Color}{\glsplain{MCDR}}}, where \textbf{\textcolor{violet}{purple}} masks denote \textit{simplified} text and \textbf{\textcolor{green}{green}} text indicates \textit{augmented} descriptions. The fifth column shows \textbf{\textcolor{orange}{3D bounding boxes}} predicted by \textbf{\textcolor{Ground_Color} {\glsplain{LLM-Grounder}}}. Visualization of each step is provided in Appendix Fig.~\ref{fig:pipeline_visualization}.}
     \label{fig:qualitative}
     ~\vspace{-0.5cm}
 \end{figure*}
 
\paragraph{\textbf{Nr3D and Sr3D datasets}.}
As shown in Tab. ~\ref{tab:table_2}, on the challenging \emph{Hard} subset -  where the target object is distracted by other objects of the same class - our approach achieves 68.8\% and 76.5\% on Nr3D and Sr3D respectively. As a result, we establish a new \textbf{state of the art in 9 out of 10 settings}, with \textit{Overall} accuracies of 75.1\% and 81.5\%, respectively.

\paragraph{\textbf{Qualitative analysis}.} As shown in the qualitative examples (Fig.~\ref{fig:qualitative}), \textbf{\textcolor{Prune_Color}{\glsplain{LGSP}}} effectively prunes irrelevant regions, reducing candidate objects and directing attention to language-relevant areas. Meanwhile, \textbf{\textcolor{DescriptReform_Color}{\glsplain{MCDR}}} rewrites descriptions into a simpler, more structured form, enriching them with additional attributes and spatial relations-even beyond the original input-while maintaining consistent length and semantic density. Together, these components enable \textbf{\textcolor{Ground_Color}{\glsplain{LLM-Grounder}}} to better align language with the target object for accurate bounding box prediction without being distracted by irrelevant candidates or missing spatial relationships in the original utterance.
 
\subsection{Ablation Study}

\begin{table}[h]
\centering
\tiny
\setlength{\tabcolsep}{2pt}
\begin{tabular}{l | ll | ll | ll}
\toprule
\multirow{2}{*}{\textbf{Method}} 
& \multicolumn{2}{c|}{\textbf{Unique}} 
& \multicolumn{2}{c|}{\textbf{Multiple}} 
& \multicolumn{2}{c}{\textbf{Overall}} \\ 
\cmidrule(lr){2-3} \cmidrule(lr){4-5} \cmidrule(lr){6-7}
& Acc\texttt{@}0.25 & Acc\texttt{@}0.5 
& Acc\texttt{@}0.25 & Acc\texttt{@}0.5 
& Acc\texttt{@}0.25 & Acc\texttt{@}0.5 \\ 
\midrule

\textbf{\textcolor{Ground_Color}{\glsplain{LLM-Grounder}}} 
& 83.1 & 75.3 
& 51.2 & 42.9 
& 57.3 & 49.1 \\

+ \textbf{\textcolor{Prune_Color}{\glsplain{LGSP}}} 
& 85.7\,{\scalebox{0.8}{\textcolor{custom_darkgreen}{+\,2.6}}} 
& 79.9\,{\scalebox{0.8}{\textcolor{custom_darkgreen}{+\,4.6}}} 
& 53.6\,{\scalebox{0.8}{\textcolor{custom_darkgreen}{+\,2.4}}} 
& 45.8\,{\scalebox{0.8}{\textcolor{custom_darkgreen}{+\,2.9}}} 
& 59.2\,{\scalebox{0.8}{\textcolor{custom_darkgreen}{+\,1.9}}} 
& 52.3\,{\scalebox{0.8}{\textcolor{custom_darkgreen}{+\,3.2}}} \\

\quad + \textbf{\textcolor{DescriptReform_Color}{\glsplain{MCDR}}}
 (S) 
& 88.6\,{\scalebox{0.8}{\textcolor{custom_darkgreen}{+\,2.9}}} 
& 82.5\,{\scalebox{0.8}{\textcolor{custom_darkgreen}{+\,2.6}}} 
& 56.0\,{\scalebox{0.8}{\textcolor{custom_darkgreen}{+\,2.4}}} 
& 47.3\,{\scalebox{0.8}{\textcolor{custom_darkgreen}{+\,1.5}}} 
& 62.4\,{\scalebox{0.8}{\textcolor{custom_darkgreen}{+\,3.2}}} 
& 54.0\,{\scalebox{0.8}{\textcolor{custom_darkgreen}{+\,1.7}}} \\

\quad + \textbf{\textcolor{DescriptReform_Color}{\glsplain{MCDR}}} (S + A) 
& \textbf{90.2}\,{\scalebox{0.8}{\textcolor{custom_darkgreen}{+\,1.6}}} 
& \textbf{84.7}\,{\scalebox{0.8}{\textcolor{custom_darkgreen}{+\,2.2}}} 
& \textbf{57.6}\,{\scalebox{0.8}{\textcolor{custom_darkgreen}{+\,1.6}}} 
& \textbf{49.4}\,{\scalebox{0.8}{\textcolor{custom_darkgreen}{+\,2.1}}} 
& \textbf{63.8}\,{\scalebox{0.8}{\textcolor{custom_darkgreen}{+\,1.4}}} 
& \textbf{56.1}\,{\scalebox{0.8}{\textcolor{custom_darkgreen}{+\,2.1}}} \\

\bottomrule
\end{tabular}
\caption{
\textbf{Impact of each proposal module on ScanRefer}.  
Baseline is \textbf{\textcolor{Ground_Color}{\glsplain{LLM-Grounder}}} trained on \textit{original} descriptions and \textit{unpruned} scenes, then incrementally add \textbf{\textcolor{Prune_Color}{\glsplain{LGSP}}} for scene pruning and \textbf{\textcolor{DescriptReform_Color}{\glsplain{MCDR}}} for description reformulation (via simplification (S) or simplification + augmentation (S + A)). Consistent improvements are also seen on Nr3D and Sr3D (see Appendix Tab. \ref{tab:ablation_2}).}
\label{tab:ablation_1}
\end{table}

\paragraph{\textbf{Impact of each proposal module.}}
As shown in Tab.~\ref{tab:ablation_1}, incorporating \textbf{\textcolor{Prune_Color}{\glsplain{LGSP}}} consistently improves performance across both \emph{Unique} and \emph{Multiple} subsets by removing irrelevant regions, yielding a \textbf{\textcolor{custom_darkgreen}{+3.2\%}} Acc\texttt{@}0.5 gain in overall accuracy. Building on this, \textbf{\textcolor{DescriptReform_Color}{\glsplain{MCDR}}} further boosts results through description reformulation: simplification alone provides noticeable improvements across both subsets (\textit{Overall} \textbf{\textcolor{custom_darkgreen}{+1.7\%}} Acc\texttt{@}0.5), while combining simplification and augmentation delivers the best performance, bringing an additional \textbf{\textcolor{custom_darkgreen}{+2.1\%}} Acc\texttt{@}0.5 gain in overall.

\paragraph{\textbf{Candidate Reduction Rate.}} Next, we conduct an ablation study to validate the effectiveness of the proposed \textbf{\textcolor{Prune_Color}{\glsplain{LGSP}}} module. We first evaluate it on the Sr3D Hard subset, where each query contains at least two distractor objects from the same category, providing a challenging setting for candidate reduction before grounding. A candidate is considered removed if the pruned scene contains no spatial region belonging to that object. As shown in Table~\ref{tab:candidate_reduction}, our module reduces grounding candidates from 2.43 to 1.28 on Sr3D Hard and from 5.72 to 1.62 on the ScanRefer Multiple split, effectively narrowing the grounding search space.

\begin{table}[h]
\centering
\small
\begin{tabular}{l | c | c}
\toprule
\textbf{Dataset} & \textbf{\makecell{No pruning}}& \textbf{\makecell{With \textbf{\textcolor{Prune_Color}{\glsplain{LGSP}}}}} \\
\midrule
ScanRefer Multiple Eval 
& 5.72 & 1.62 \\
Sr3D Hard set
& 2.43 & 1.28 \\
\bottomrule
\end{tabular}
\caption{\textbf{Impact of} \textbf{\textcolor{Prune_Color}{\glsplain{LGSP}}} \textbf{on removing noisy candidates in cluttered environments.}. The average number of candidates is compared against no-pruning method.}
\label{tab:candidate_reduction}
\end{table}
\begin{table}[t]
\centering
\small
\begin{tabular}{c | ll}
\toprule
\textbf{\# Oblique views} 
& \textbf{RGB} 
& \textbf{RGB + Depth} \\
\midrule

1 
& 72.6 
& 74.3\,{\scalebox{0.8}{\textcolor{custom_darkgreen}{+\,1.7}}} \\

2 
& 84.8\,{\scalebox{0.8}{\textcolor{custom_darkgreen}{+\,12.2}}} 
& 88.9\,{\scalebox{0.8}{\textcolor{custom_darkgreen}{+\,16.3}}} \\

4 
& 93.1\,{\scalebox{0.8}{\textcolor{custom_darkgreen}{+\,20.5}}} 
& \textbf{94.7}\,{\scalebox{0.8}{\textcolor{custom_darkgreen}{+\,22.1}}} \\

\bottomrule
\end{tabular}
\caption{
\textbf{Target recall of \textbf{\textcolor{Prune_Color}{\glsplain{LGSP}}} under varying numbers of oblique views} on the ScanRefer dataset.
A sample is counted as successful when the pruned region retains $>$75\% of the target, 
and target recall is reported as the percentage of successful samples. 
Improvements are measured against the 1-view RGB baseline (72.6)}
\label{tab:view}
\end{table}
\paragraph{\textbf{Target recall of LGSP under varying numbers of oblique views.}}
In Tab.~\ref{tab:view}, we evaluate how the number of oblique views affects relevant-region detection, where success is defined as covering $>$75\% of the target. Performance consistently improves with more views and further benefits from depth. Notably, using two opposing oblique views with depth already yields strong gains, while four views achieve the best result of 94.7\% (\textbf{\textcolor{custom_darkgreen}{+22.1\%}} over the 1-view RGB baseline). Based on this, we adopt four oblique views with depth as a good trade-off between accuracy and efficiency, with optional box expansion to increase target coverage in practice.

\begin{table}[h]
\centering
\small
\setlength{\tabcolsep}{2pt}
\begin{tabular}{l|lll}
\toprule
\textbf{Frozen VLMs} & \makecell{\textbf{Area}\\\textbf{retained}\\\textbf{(\%)}}& \makecell{\textbf{Target}\\\textbf{recall}\\\textbf{(\%)}} & \makecell{\textbf{Run time}\\\textbf{(second)}} \\
\midrule

Gemini 3 Pro
& 12.1 & \textbf{95.7}& 13.1 \\
ChatGPT-5
& 11.7 & 95.3 & 11.2 \\
Qwen2.5-VL-72B-Instruct
& 16.3 & 95.1 & 12.6 \\
\rowcolor{custom_light_purple_2}
Qwen2.5-VL-3B-Instruct
& 15.1 & 94.7\,{\scalebox{0.8}{\textcolor{red}{-\,1.0}}} & \textbf{1.4}\,{\scalebox{0.8}{\textcolor{custom_darkgreen}{-\,11.7}}} \\

\bottomrule
\end{tabular}
\caption{\textbf{Analysis of different VLMs on the pruning task}, used in \textbf{\textcolor{Prune_Color}{\glsplain{LGSP}}} and \textbf{\textcolor{DescriptReform_Color}{\glsplain{MCDR}}}. Box sizes are measured in pixels with respect to a 512$\times$512 top-view image. Deviations
are measured against the best results. The entire framework using Qwen2.5-VL-3B-Instruct needs 4.1s for inference (see Appendix  Tab.~\ref{tab:running_time}).}
\label{tab:vlms}
\end{table}

\paragraph{\textbf{VLM selection justification}} 
While larger VLMs such as Qwen2.5-VL-72B, used in SeeGround~\cite{Li2024SeeGroundSA}, and frontier APIs such as ChatGPT-5 and Gemini 3 Pro offer stronger general-purpose reasoning, they incur substantial latency ($>$10s/sample) and cost, limiting large-scale evaluation. As shown in Tab.~\ref{tab:vlms}, we compare four VLMs with a \textit{no-pruning} baseline, which retains the full $512\times512$ top-view image, across pruned area, target recall, and runtime. All four VLMs achieve similar recall ($94.7$--$95.7\%$) while removing $83.7$--$88.3\%$ of irrelevant regions, suggesting that the pruning task is largely saturated and gains little from frontier-scale reasoning. Qwen2.5-VL-3B-Instruct~\cite{Bai2025Qwen25VLTR} runs in $1.4$s per stage, about $8$--$9\times$ faster than the alternatives, while remaining within $1\%$ recall of the best model.

\section{Conclusion}

In this paper, we introduce \PruneGround, the first framework to leverage spatial pruning and detection-pretrained spatial \glsplain{LLM} repurposing for \glsplain{3DVG}. It integrates three key components: (i) \textbf{\textcolor{Prune_Color}{\glsplain{LGSP}}} to remove irrelevant regions using a frozen \glsplain{VLM} with semantic and geometric cues, (ii) \textbf{\textcolor{DescriptReform_Color}{\glsplain{MCDR}}} to reformulate the description into a simpler form while augmenting it with semantic and relational spatial information, and (iii) \textbf{\textcolor{Ground_Color}{\glsplain{LLM-Grounder}}}, an \glsplain{LLM}-based grounding model that aligns point cloud features with linguistic representations for target-selective prediction. Our \textit{plug-and-play} framework achieves \textbf{state-of-the-art} results on three most popular point cloud benchmarks (ScanRefer, Nr3D and Sr3D). Notably, it demonstrates strong generalization by delivering state-of-the-art results on all three ScanRefer settings and on 9 out of 10 Nr3D/Sr3D settings.

\section*{Limitations}
Our pipeline is limited by \textit{error propagation} across its three stages. Errors in \textbf{\textcolor{Prune_Color}{\glsplain{LGSP}}} can produce incomplete or noisy candidates, while mistakes in language understanding in \textbf{\textcolor{DescriptReform_Color}{\glsplain{MCDR}}} may misinterpret attributes or spatial relations. Since  \textbf{\textcolor{Ground_Color}{\glsplain{LLM-Grounder}}} relies on these intermediate outputs, early-stage errors can accumulate and amplify, reducing robustness in cluttered scenes, ambiguous descriptions, and fine-grained spatial reasoning tasks (see Fig. \ref{fig:failure_case_analysis} of Appendix Sec. \ref{sec:failure_case_analysis}).

Furthermore, future work should focus on developing more lightweight and adaptable approaches to improve generalization across diverse real-world environments.

\section*{Acknowledgement}
This work was supported in part by the German Federal Ministry of Education and Research (BMBF): Tübingen AI Center, FKZ: 01IS18039B; by the Machine Learning Cluster of Excellence, EXC number 2064/1 – Project number 390727645; by Coefficient Giving; by the SwissAI Initiative; by Schmidt Sciences; by the Frontier Model Forum and AI Safety Fund;by the Canadian AI Safety Institute Research Program at CIFAR; by the Canadian AI Safety Institute Research Program at CIFAR through a Catalyst Award; by the Survival and Flourishing Fund; by the Cooperative AI Foundation; and by the Division of Engineering Science at the University of Toronto. Resources used in preparing this research project were provided, in part, by the Province of Ontario, the Government of Canada through CIFAR, and companies sponsoring the Vector Institute.
F.D.
acknowledges support through a fellowship from the Hector
Fellow Academy.

\bibliography{ms}

\clearpage 


\appendix

\onecolumn
\begin{center}
    {\Large\textsf{\PruneGround \, Supplementary}}
\end{center}

\tableofcontents
\addtocontents{toc}{\protect\setcounter{tocdepth}{2}}

\onecolumn
\section{Cognitive Motivation}
\label{sec:cognitive_motivation}
Our motivation is also consistent with evidence from human visual cognition. Humans rarely interpret a complex scene by exhaustively comparing all objects of the same category. Instead, visual search is guided by selective attention \cite{le2026vision, treisman1980feature, wolfe1994guided}: early perceptual processing extracts broad cues in parallel, while more complex object identification and relational reasoning are applied only to a limited set of task-relevant regions. In situated language understanding, linguistic cues further modulate this attentional process. Eye-tracking studies show that listeners rapidly direct their gaze toward objects mentioned or implied by an unfolding utterance, indicating that language incrementally guides visual attention in the surrounding environment \cite{tanenhaus1995integration, huettig2011using}. 

This cognitive process is especially relevant to \glsplain{3DVG}. Spatial language typically describes a target object, or located object, with respect to one or more reference objects, such as "the chair closest to the door" or "the chair under chalkboard". Cognitive studies of spatial language suggest that understanding such expressions involves identifying the relevant objects, assigning a reference frame, and evaluating the spatial relation between the target and the anchor object rather than globally reasoning over every possible candidate in the scene \cite{logan1994spatial, carlson2001using}.

Inspired by this human-like strategy, our \PruneGround \, avoids treating all object candidates as equally relevant. Given a referring expression, it first uses the mentioned anchors and spatial relations to narrow the search space to a language-conditioned local region (\textbf{\textcolor{Prune_Color}{\glsplain{LGSP}}}). The model then performs fine-grained grounding within this region, augmenting with candidates that are spatially and semantically plausible (\textbf{\textcolor{DescriptReform_Color}{\glsplain{MCDR}}}). This design mirrors how humans use linguistic context to guide attention toward informative scene regions, reducing unnecessary comparison with irrelevant distractors while preserving the relational cues needed for accurate localization. Consequently, the localization (\textbf{\textcolor{Ground_Color}{\glsplain{LLM-Grounder}}}) provides a cognitively grounded alternative to exhaustive \textit{segment-then-select} pipelines, improving both accuracy and robustness in cluttered 3D environments.

\onecolumn
\section{Camera Setup Visualization}
\label{sec:camera_setup}
\begin{figure}[h]
    \centering
    \includegraphics[width=0.8\linewidth]{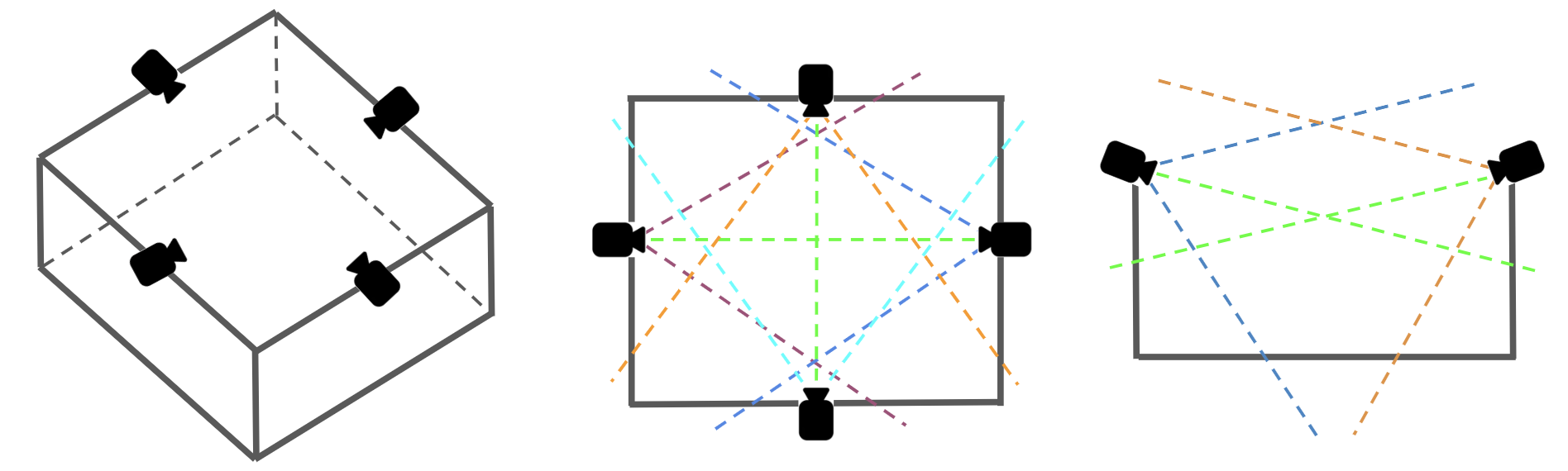}
    \caption{\textbf{Camera setup visualization: \textit{Global View Rendering}}. From left to right: Cameras with 3D object bounding boxes, top-view visualization, and side-view visualization.}
    \label{fig:sup_global_cam}
\end{figure}

\paragraph{Global View Rendering.} We provide a visualization of the camera setup for \textit{Global View Rendering} - a part of \textbf{\textcolor{Prune_Color}{\glsplain{LGSP}}} in Sec. \ref{LGSP}. As shown in Fig.~\ref{fig:sup_global_cam}, four cameras are placed at the centers of the four upper edges of the 3D bounding box, each oriented toward the center of the opposite face.

\begin{figure}[h]
    \centering
    \includegraphics[width=\linewidth]{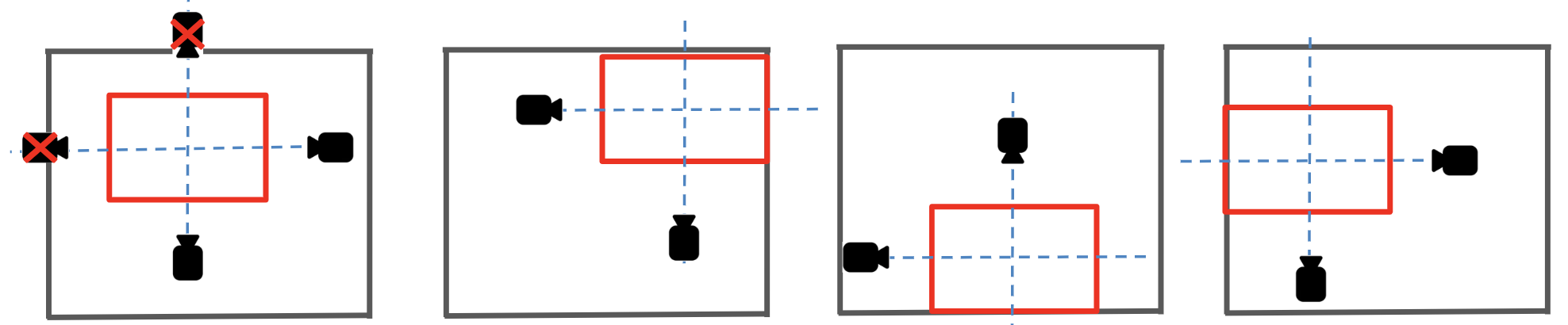}
    \caption{\textbf{Camera setup visualization: \textit{Local View Rendering}} in four scenarios.}
    \label{fig:sup_local_cam}
\end{figure}

\paragraph{Local View Rendering.} We provide a visualization of the camera setup for \textit{Local View Rendering} - a part of \textbf{\textcolor{DescriptReform_Color}{\glsplain{MCDR}}} in Sec. \ref{MCDR}. Given a 3D bounding box, we render side views in both the horizontal and vertical directions. As shown in the first example in Fig.~\ref{fig:sup_local_cam}, for each direction there are two possible camera positions that can cover the box; however, we select the one closer to the center of the original scene (the center of the black box). Note that the camera is then slightly elevated to obtain a clearer view and to prevent occlusion by surrounding furniture.

\onecolumn
\section{VLM Query Structure}
\label{sec:VLM_query_prompt}
In this section, we describe the query structure, which consists of two tasks.
\paragraph{\textbf{Prompt for 2D Bounding Box Estimation.}}
The Fig. \ref{fig:prompt_2d_bbox_estimation} shows the prompt for 2D bounding box estimation. Given the input description, the scene size $(m, n, k)$, and the top-view image resolution $(w, h)$, the objective is to estimate the corresponding 2D bounding box on the top-view image. This prompt is used for \textbf{Language-Guided Region Selection} in \textbf{\textcolor{Prune_Color}{\glsplain{LGSP}}} (Sec. \ref{LGSP}).

\begin{tcolorbox}[
    breakable,
    enhanced,
    colback=pink!3, 
    colframe=Ground_Color, 
    coltitle=violet,
    title={\PruneGround \,Prompt for \newline 2D Bounding Box Estimation},
    fonttitle=\bfseries,
    sharp corners,
    boxrule=0.8pt,
    colbacktitle=purple!5, 
    attach boxed title to top left={xshift=2mm,yshift=-2mm}
]

\begin{lstlisting}[style=mypython, language=Python]
messages = [
    {
        "role": "user",
        "content": [

            {"type": "image", "image": "topview.png"},

            {"type": "image", "image": "view_1_rgb.png"},
            {"type": "image", "image": "view_1_depth.png"},

            {"type": "image", "image": "view_2_rgb.png"},
            {"type": "image", "image": "view_2_depth.png"},

            {"type": "image", "image": "view_3_rgb.png"},
            {"type": "image", "image": "view_3_depth.png"},

            {"type": "image", "image": "view_4_rgb.png"},
            {"type": "image", "image": "view_4_depth.png"},

            {
                "type": "text",
                "text": f"""
Task:
Given the description: "{description}", Determine the relevant region in the topview image as a 2D bounding box.

Image order:
The images appear in the following order:
1. topview.png  
2. view_1_rgb.png  
3. view_1_depth.png  
4. view_2_rgb.png  
5. view_2_depth.png  
6. view_3_rgb.png  
7. view_3_depth.png  
8. view_4_rgb.png  
9. view_4_depth.png  

Scene coordinate system:
- X axis: horizontal width = {m} meters
- Y axis: vertical length = {n} meters
- Z axis: upward height = {k} meters

Topview image:
- topview.png is a topview rendering of the scene
- width = {w} pixels corresponding to the X dimension
- height = {h} pixels corresponding to the Y dimension

Image coordinate system:
- (0, 0) is the top-left corner of the topview image
- x increases to the right
- y increases downward

Oblique views:
- Each oblique view consists of two consecutive images:
first the RGB image, then the corresponding depth image.
- The cameras are placed at the midpoint of the four top 
edges of the axis-aligned 3D bounding box.

View mapping:
- view_1 (images 2-3: view_1_rgb.png, view_1_depth.png)  
  camera looks from bottom to top in the topview
- view_2 (images 4-5: view_2_rgb.png, view_2_depth.png)  
  camera looks from right to left in the topview
- view_3 (images 6-7: view_3_rgb.png, view_3_depth.png)  
  camera looks from top to bottom in the topview
- view_4 (images 8-9: view_4_rgb.png, view_4_depth.png) 
  camera looks from left to right in the topview

Reasoning procedure:
Step 1: Identify possible target objects and anchors mentioned in the description.  
Step 2: Use the topview and oblique RGB-depth images to determine plausible spatial regions containing target candidates.  
Step 3: Determine the minimal bounding region that covers those candidates.

Bounding box requirements:
- cover all candidate regions that could plausibly contain the target object
- exclude regions that are unlikely to contain the target
- include anchors mentioned in the description only when they are spatially close to the target candidates
- if the anchors are far from the target candidates, they should not be included
- the bounding box should be as tight as possible while still covering all plausible target candidates
- the bounding box must not exceed the scene boundary and should avoid covering large empty areas

If you are uncertain, prefer a slightly larger box that still excludes clearly irrelevant regions.

Output format:
- Return ONLY four values: x_min, x_max, y_min, y_max
- All coordinates must be in pixel coordinates of the topview image.
- Output ONLY the four numbers separated by commas.
- Do not output any explanation or additional text.
"""
            },
        ],
    }
]
\end{lstlisting}
\end{tcolorbox}
\captionof{figure}{\textbf{Prompt for 2D bounding box estimation: Top-view visual grounding.}}
\label{fig:prompt_2d_bbox_estimation}

\onecolumn
\paragraph{Prompt for \textbf{\textcolor{DescriptReform_Color}{\glsplain{MCDR}}}.}
The \glsplain{VLM} query for \textbf{Description Simplification and Augmentation} - a part of \textbf{\textcolor{DescriptReform_Color}{\glsplain{MCDR}}} (Sec. \ref{MCDR}) is described in Fig. \ref{fig:prompt_MCDR}. Given rendered local multi-view images  and the original description, we prompt the \glsplain{VLM} to augment the relational information by using the local multi-view images, thereby incorporating not only reformulated content from the original description but also additional target--anchor pairs.

\begin{tcolorbox}[
    breakable,
    enhanced,
    colback=pink!3, 
    colframe=Ground_Color, 
    coltitle=violet,
    title=\PruneGround \,Prompt for \textbf{\textcolor{DescriptReform_Color}{\glsplain{MCDR}}},
    fonttitle=\bfseries,
    sharp corners,
    boxrule=0.8pt,
    colbacktitle=purple!5, 
    attach boxed title to top left={xshift=2mm,yshift=-2mm}
]
\begin{lstlisting}[style=mypython, language=Python]
messages = [
    {
        "role": "user",
        "content": [

            {"type": "image", "image": "side_1.png"},
            {"type": "image", "image": "side_2.png"},

            {
                "type": "text",
                "text": f"""
Task:

You are given two side-view images of a pruned 3D scene.

Images:
1. side_1.png
2. side_2.png

Description: "{description}"
Your goal is to analyze the description and the two side views to extract the target object and anchors, and reformulate the description into structured single target-single anchor utterances.

Reasoning procedure:
Step 1 -  Analysis
Analyze the description and side views to identify:
- the target object
- the list of anchors explicitly mentioned
in the description
Step 2 - Visual Anchor Detection
From the two side-view images, identify additional anchors that are visible in the scene but NOT mentioned in the description.
Step 3 - Reformulate Description
Reformulate the description into a list of single target - single anchor utterances.
Each utterance should describe the spatial relation between the target and one anchor.

Output Structure:
- target name
- anchors name:
[list of anchors mentioned in the description]
- anchors not in description but visible:
[list of anchors visible in side views but not mentioned in the description]
- reformulated_description:
    + target properties
    (short and concise description of the target's properties)
    + augmented target properties
    (additional target properties that can be inferred
    from the side views but are not explicitly described)
    + single target single anchor utterances
    (list of utterances where each describes the relation between the target and one anchor)
    + augmented single target - single anchor utterances
    (list of utterances using anchors that are not mentioned in the description but detected in the side views)

Constraints:
- target properties and augmented target properties must be short and concise.
- the TOTAL number of single target - single anchor utterances must be at most 5.
- each utterance must follow this format:
target (+properties) + spatial relation + anchor (+properties)
- Use only clear spatial relations such as:
left of, right of, in front of, behind, next to, under, on, ...
- Do not produce explanations.
- Return only the structured output.

Example target properties:
- the chair is wooden and yellow
Example utterance:
- the chair is behind the black coffee table
"""
            },
        ],
    }
]
\end{lstlisting}
\end{tcolorbox}
\captionof{figure}{\textbf{Prompt for \textbf{\textcolor{DescriptReform_Color}{\glsplain{MCDR}}}.}}
\label{fig:prompt_MCDR}

\onecolumn
\section{Additional Ablation Study}

\paragraph{Inference analysis.}
In this study, we report the runtime and resource consumption of \PruneGround \, during inference. We evaluate our pipeline on GPU RTX A100, reported by Tab.~\ref{tab:running_time}.

\begin{table}[h]
\centering
\small
\begin{tabular}{l | c | c}
\toprule
Module & Runtime & GPU usage \\ 
\midrule
\textbf{\textcolor{Prune_Color}{\glsplain{LGSP}}} & 1.4 & 10 \\
\textbf{\textcolor{DescriptReform_Color}{\glsplain{MCDR}}} & 1.3 & 8 \\ 
\textbf{\textcolor{Ground_Color}{\glsplain{LLM-Grounder}}} & 1.4 & 5\\
\rowcolor{custom_light_purple_2}
\PruneGround & 4.1 (total) & 10 (max)\\ 
\bottomrule
\end{tabular}
\caption{\textbf{Inference analysis} based on running time (in seconds) and GPU usage (in GBs) of our framework.The frozen \glsplain{VLM} used in \textbf{\textcolor{Prune_Color}{\glsplain{LGSP}}} and \textbf{\textcolor{DescriptReform_Color}{\glsplain{MCDR}}} is Qwen2.5-VL-3B-Instruct. Its inference time comparison against other \glsplain{VLM}s is shown in Tab. \ref{tab:vlms}.}
\label{tab:running_time}
\end{table}

\paragraph{Performance with other grounders.}
Since our pruning mechanism is designed as a \textit{plug-and-play} module, the pruned regions can be directly used as input to other grounders. To verify that reducing the search space improves grounding performance, we retrain ScanRefer~\cite{chen2020scanrefer} and ReferIt3D~\cite{Achlioptas2020ReferIt3DNL}, which are foundational works that introduced both the methods and datasets for the \glsplain{3DVG} task. Both methods follow the \textit{segment-then-select} paradigm. As shown in Tab.~\ref{tab:other_grounders}, integrating our language-based pruning consistently improves their performance, demonstrating its effectiveness for \glsplain{3DVG}. Additionally, we apply our pruning module to 3DVG-Transformer~\cite{Zhao20213DVGTransformerRM}, a representative \textit{one-stage} approach, and observe consistent improvements.
\begin{table*}[h]
\centering
\small
\begin{tabular}{l | ccccc}
\toprule
Grounder  & ScanRefer Acc\texttt{@}0.25 & ScanRefer Acc\texttt{@}0.5 & Nr3D Overall & Sr3D Overall \\ 
\midrule
ScanRefer (\textit{two-stage})  & 37.30 &  24.32&-&-\\
ScanRefer + LGSP & \textbf{44.53}\,{\scalebox{0.8}{\textcolor{custom_darkgreen}{+\,7.23}}}  
& \textbf{28.46}\,{\scalebox{0.8}{\textcolor{custom_darkgreen}{+\,4.14}}} &-&-\\
\midrule
ReferIt3D (\textit{two-stage}) &-&-&35.6&40.8\\
ReferIt3D + LGSP &-&-&\textbf{43.8}\,{\scalebox{0.8}{\textcolor{custom_darkgreen}{+\,3.2}}} 
&\textbf{44.6}\,{\scalebox{0.8}{\textcolor{custom_darkgreen}{+\,3.8}}} \\
\midrule
3DVG-Transformer (\textit{one-stage}) & 45.90 & 34.47 & 40.8 & 51.4\\
3DVG-Transformer + LGSP 
& \textbf{50.25}\,{\scalebox{0.8}{\textcolor{custom_darkgreen}{+\,4.35}}} 
& \textbf{37.52}\,{\scalebox{0.8}{\textcolor{custom_darkgreen}{+\,3.05}}} 
& \textbf{45.3}\,{\scalebox{0.8}{\textcolor{custom_darkgreen}{+\,4.5}}} 
& \textbf{55.6}\,{\scalebox{0.8}{\textcolor{custom_darkgreen}{+\,4.2}}} \\
\bottomrule
\end{tabular}
\caption{\textbf{Performance of \textbf{\textcolor{Prune_Color}{\glsplain{LGSP}}} and \textbf{\textcolor{DescriptReform_Color}{\glsplain{MCDR}}} when plug-and-played with different grounding methods.}}
\label{tab:other_grounders}
\end{table*}
\paragraph{Ablation study of proposed modules on Nr3D and Sr3D.}
We further provide an ablation study of each proposed module on the Nr3D and Sr3D datasets in Tab.~\ref{tab:ablation_2}, complementing the study on ScanRefer dataset in Tab.~\ref{tab:ablation_1}. Consistent improvements are observed across both datasets, confirming the effectiveness of the proposed modules.

\begin{table*}[h]
\centering
\small
\setlength{\tabcolsep}{2pt}
\begin{tabular}{l | lllll | lllll}
\toprule
\multirow{1}{*}{\textbf{Method}} 
& \multicolumn{5}{c|}{\textbf{Nr3D}} 
& \multicolumn{5}{c}{\textbf{Sr3D}} \\
\cmidrule(lr){2-6} \cmidrule(lr){7-11}
 &  Easy & Hard & View-D & View-I & Overall 
 & Easy & Hard & View-D & View-I & Overall \\
\midrule

\textbf{\textcolor{Ground_Color}{\glsplain{LLM-Grounder}}} 
& 79.8 & 61.7 & 66.2 & 72.1 & 69.7 
& 77.0 & 69.7 & 56.1 & 75.6 & 74.8 \\

+ \textbf{\textcolor{Prune_Color}{\glsplain{LGSP}}} 
& 82.1\,{\scalebox{0.8}{\textcolor{custom_darkgreen}{+\,2.3}}} 
& 64.6\,{\scalebox{0.8}{\textcolor{custom_darkgreen}{+\,2.9}}} 
& 68.4\,{\scalebox{0.8}{\textcolor{custom_darkgreen}{+\,2.2}}} 
& 74.3\,{\scalebox{0.8}{\textcolor{custom_darkgreen}{+\,2.2}}}
& 72.1\,{\scalebox{0.8}{\textcolor{custom_darkgreen}{+\,2.4}}} 

& 79.2\,{\scalebox{0.8}{\textcolor{custom_darkgreen}{+\,2.2}}} 
&  72.9\,{\scalebox{0.8}{\textcolor{custom_darkgreen}{+\,3.2}}}
& 58.6\,{\scalebox{0.8}{\textcolor{custom_darkgreen}{+\,2.5}}} 
& 78.1\,{\scalebox{0.8}{\textcolor{custom_darkgreen}{+\,2.5}}} 
& 77.3\,{\scalebox{0.8}{\textcolor{custom_darkgreen}{+\,2.5}}} \\

\quad + \textbf{\textcolor{DescriptReform_Color}{\glsplain{MCDR}}}
& 83.5\,{\scalebox{0.8}{\textcolor{custom_darkgreen}{+\,2.2}}}
& 66.8\,{\scalebox{0.8}{\textcolor{custom_darkgreen}{+\,2.2}}}
& 69.8\,{\scalebox{0.8}{\textcolor{custom_darkgreen}{+\,1.4}}}
& 75.3\,{\scalebox{0.8}{\textcolor{custom_darkgreen}{+\,1.0}}}
& 73.6\,{\scalebox{0.8}{\textcolor{custom_darkgreen}{+\,1.5}}}

& 80.8\,{\scalebox{0.8}{\textcolor{custom_darkgreen}{+\,1.6}}}
& 75.1\,{\scalebox{0.8}{\textcolor{custom_darkgreen}{+\,2.2}}}
& 59.6\,{\scalebox{0.8}{\textcolor{custom_darkgreen}{+\,1.0}}}
& 80.2\,{\scalebox{0.8}{\textcolor{custom_darkgreen}{+\,2.1}}}
& 79.1\,{\scalebox{0.8}{\textcolor{custom_darkgreen}{+\,1.8}}} \\

\quad + \textbf{\textcolor{DescriptReform_Color}{\glsplain{MCDR}}} (S + A)
& \textbf{84.6}\,{\scalebox{0.8}{\textcolor{custom_darkgreen}{+\,1.1}}} 
& \textbf{68.8}\,{\scalebox{0.8}{\textcolor{custom_darkgreen}{+\,2.0}}}
& \textbf{71.2}\,{\scalebox{0.8}{\textcolor{custom_darkgreen}{+\,1.4}}}
& \textbf{76.6}\,{\scalebox{0.8}{\textcolor{custom_darkgreen}{+\,1.3}}}
& \textbf{75.1}\,{\scalebox{0.8}{\textcolor{custom_darkgreen}{+\,1.5}}}

& \textbf{83.7}\,{\scalebox{0.8}{\textcolor{custom_darkgreen}{+\,2.9}}}
& \textbf{76.5}\,{\scalebox{0.8}{\textcolor{custom_darkgreen}{+\,1.4}}}
& \textbf{61.7}\,{\scalebox{0.8}{\textcolor{custom_darkgreen}{+\,2.1}}}
& \textbf{82.4}\,{\scalebox{0.8}{\textcolor{custom_darkgreen}{+\,2.2}}}
& \textbf{81.5}\,{\scalebox{0.8}{\textcolor{custom_darkgreen}{+\,2.4}}} \\

\bottomrule
\end{tabular}
\caption{
\textbf{Impact of each proposal module on Nr3D and Sr3D datasets}.  
Baseline is \textbf{\textcolor{Ground_Color}{\glsplain{LLM-Grounder}}} trained on \textit{original} descriptions and \textit{unpruned} scenes, then incrementally add \textbf{\textcolor{Prune_Color}{\glsplain{LGSP}}} for scene pruning and \textbf{\textcolor{DescriptReform_Color}{\glsplain{MCDR}}} for description reformulation (via simplification (S) or simplification + augmentation (S + A)).}

\label{tab:ablation_2}
\end{table*}
\paragraph{Anchor verification analysis.}
To assess whether \textbf{\textcolor{DescriptReform_Color}{\glsplain{MCDR}}} introduces meaningful spatial information rather than spurious content, we analyze the anchors before and after reformulation. For each augmented anchor (i.e., anchors not present in the original description but added by \textbf{\textcolor{DescriptReform_Color}{\glsplain{MCDR}}}), we verify its existence against ScanNet object annotations within the pruned region. An augmented anchor is marked as \emph{verified} if an instance of its class label actually exists in the pruned scene, and \emph{hallucinated} otherwise. We perform this analysis on 500 randomly sampled descriptions from each dataset.

As shown in Tab.~\ref{tab:anchor_verification}, \textbf{\textcolor{DescriptReform_Color}{\glsplain{MCDR}}} increases the average number of anchors per description from 2.14 (original) to 3.22 (reformulated) on ScanRefer, with similar trends on Nr3D and Sr3D. Among the augmented anchors, average 93.7\% are verified as existing in the pruned scene. The hallucination rate remains below 6.3\%, demonstrating that visual conditioning effectively constrains the VLM to ground augmentations in actual scene content rather than fabricating plausible-sounding but non-existent references. This verification analysis confirms that the additional anchors introduced by \textbf{\textcolor{DescriptReform_Color}{\glsplain{MCDR}}} are predominantly grounded in the scene and contribute genuine spatial information, supporting the downstream grounding improvements.

\begin{table}[t]
\centering
\small
\setlength{\tabcolsep}{4pt}
\begin{tabular}{lcccc}
\toprule
\textbf{Dataset} & \makecell{\textbf{\# Anch.}\\\textbf{(Orig.)}} & \makecell{\textbf{\# Anch.}\\\textbf{\textcolor{DescriptReform_Color}{\glsplain{MCDR}}}} & \makecell{\textbf{Verif.}\\(\%)} & \makecell{\textbf{Halluc.}\\(\%)} \\
\midrule
ScanRefer & 2.14 & 3.22 & 93.4 & 6.6  \\
Nr3D      & 1.07 & 2.12 & 91.7 & 8.3 \\
Sr3D      & 0.72 & 1.75 & 96.1 & 5.9 \\
\midrule
Average & 1.31 & 2.36 & 93.7 & 6.3  \\
\bottomrule
\end{tabular}
\caption{\textbf{Anchor verification} on 500 sampled descriptions per dataset. \textbf{Verif.}: augmented anchors whose class exists in the pruned region. \textbf{Halluc.}: augmented anchors with no corresponding object instance in the pruned region.}
\label{tab:anchor_verification}
\end{table}

\paragraph{Human evaluation of Reformulated Descriptions.}
To assess the linguistic quality of \textbf{\textcolor{DescriptReform_Color}{\glsplain{MCDR}}} beyond downstream grounding accuracy, we conduct a human study comparing three description variants: (i) the \emph{Original} free-form utterance, (ii) a \emph{Text-only Reformulation} produced by prompting the same VLM with the original description but no visual input, and (iii) \textbf{\textcolor{DescriptReform_Color}{\glsplain{MCDR}}} (ours), which conditions the VLM on multi-view renderings of the pruned scene.

We randomly sample 150 descriptions from ScanRefer (50), Nr3D (50), and Sr3D (50). Three annotators with backgrounds in computer vision and natural language processing independently rate each description on a 5-point scale (1 = poor, 5 = excellent) along four dimensions \cite{le2025s}:
(i) \textbf{Faithfulness} -- whether the reformulation preserves the original intent without distorting the target referent;
(ii) \textbf{Completeness} -- whether useful spatial information is added relative to the original;
(iii) \textbf{Spatial Precision} -- whether the spatial relations are accurate and specific (e.g., ``left of'' vs.\ vague ``near'');
(iv) \textbf{Conciseness} -- whether the description is compact and free of redundancy.
Annotators are provided with the corresponding 3D scene visualization and the ground-truth target bounding box. Inter-annotator agreement is measured via Krippendorff's $\alpha$.

\begin{table}[t]
\centering
\small
\setlength{\tabcolsep}{2pt}
\begin{tabular}{lcccc}
\toprule
\textbf{Method} & \makecell{\textbf{Faith.}\\$\uparrow$} & \makecell{\textbf{Compl.}\\$\uparrow$} & \makecell{\textbf{Sp. Prec.}\\$\uparrow$} & \makecell{\textbf{Concise.}\\$\uparrow$} \\
\midrule
Original                  & 4.21 & 2.84 & 3.15 & \textbf{4.52} \\
Text-only Reformulation   & 3.58 & 3.62 & 3.23 & 3.31 \\
MCDR (Ours)               & \textbf{4.47} & \textbf{4.39} & \textbf{4.28} & 3.67 \\
\midrule
Krippendorff's $\alpha$   & 0.72 & 0.69 & 0.74 & 0.61 \\
\bottomrule
\end{tabular}
\caption{\textbf{Human evaluation of description quality} (mean rating on a 5-point scale across 150 samples, 3 annotators). Bold indicates the best per column. Krippendorff's $\alpha \geq 0.67$ indicates substantial agreement across all dimensions. \textit{Text-only Reformulation} follows a similar text only restructuring philosophy to ViewSRD~\cite{Huang_2025_ICCV}, but uses our prompting protocol, the comparison thus isolates the contribution of visual conditioning in \textbf{\textcolor{DescriptReform_Color}{\glsplain{MCDR}}}.}
\label{tab:human_eval}
\end{table}

As reported in Tab.~\ref{tab:human_eval}, \textbf{\textcolor{DescriptReform_Color}{\glsplain{MCDR}}} achieves the highest scores on \textbf{Faithfulness}, \textbf{Completeness}, and \textbf{Spatial Precision}, outperforming both the Original utterance and the Text-only Reformulation. Notably, while Text-only Reformulation also enriches the original description with additional content, it underperforms \textbf{\textcolor{DescriptReform_Color}{\glsplain{MCDR}}} on Faithfulness and Spatial Precision -- a gap attributable to its lack of visual grounding, which causes the VLM to fabricate plausible-sounding but inaccurate spatial relations. In contrast, \textbf{\textcolor{DescriptReform_Color}{\glsplain{MCDR}}} leverages multi-view scene context to anchor augmentations in actual geometry, yielding spatially precise and faithful descriptions. The Conciseness score of \textbf{\textcolor{DescriptReform_Color}{\glsplain{MCDR}}} is slightly lower than the Original, since reformulation inevitably expands the utterance, however, this trade-off is justified by the substantial gains in Completeness and Spatial Precision, which translate into improved downstream grounding. These results indicate that the improvements brought by \textbf{\textcolor{DescriptReform_Color}{\glsplain{MCDR}}} are not merely an artifact of training-time supervision, but stem from genuine enhancements in the linguistic quality of the referring expressions themselves.

\onecolumn
\section{Model Structure and Training Procedure}
\label{sup: training}
\paragraph{Qwen2.5-VL-3B-Instruct.} We use this frozen \glsplain{VLM} for inference inside \textbf{\textcolor{Prune_Color}{\glsplain{LGSP}}} and \textbf{\textcolor{DescriptReform_Color}{\glsplain{MCDR}}}. The language backbone consists of 36 transformer layers \cite{vaswani2017attention} with a hidden size of 2048, SiLU activations \cite{elfwing2018sigmoid}, RMS normalization \cite{zhang2019root}, 16 attention heads, and grouped-query attention with 2 key--value heads. The model uses a vocabulary of 151{,}936 tokens, tied input--output embeddings, bfloat16 precision, rotary positional embeddings with \(\theta_{\mathrm{RoPE}} = 1{,}000{,}000\), and mRoPE scaling for multimodal position encoding, supporting a maximum context length of 128{,}000 tokens. Although the configuration specifies a sliding-window size of 32{,}768 tokens, sliding-window attention was disabled during inference. The visual encoder contains 32 layers \cite{dosovitskiyimage} with hidden size 1280, 16 attention heads, \(14 \times 14\) spatial patches, spatial merge size 2, and selected full-attention blocks at layers 7, 15, 23, and 31; its output is projected to the 2048-dimensional language-model hidden space. We used the model strictly for inference, with caching enabled and without any parameter updates or additional training.
\paragraph{SpatialLM.} As \textbf{\textcolor{Ground_Color}{\glsplain{LLM-Grounder}}}, we use \textbf{SpatialLM1.1-Qwen-0.5B}, a spatial-language causal decoder model. SpatialLM is designed with an Encoder--MLP--LLM architecture. We use Qwen2.5-0.5B \cite{qwen2} as the base LLM, Sonata \cite{wu2025sonata} as the point-cloud encoder, and a two-layer MLP as the projector. The language backbone follows a Qwen-style transformer architecture \cite{vaswani2017attention} with 24 hidden layers, hidden size 896, intermediate size 4864, SiLU activations \cite{elfwing2018sigmoid}, RMS normalization with $\epsilon=10^{-6}$ \cite{zhang2019root}, and 14 attention heads with grouped-query attention using 2 key-value heads. The model uses a vocabulary of 151{,}936 tokens with tied input-output embeddings, rotary positional embeddings with \texttt{rope\_theta}$=1{,}000{,}000$, a maximum context length of 32{,}768 tokens, and no sliding-window attention. Spatial inputs are represented using dedicated point start, end, and point tokens, and encoded with a Sonata point backbone. The point encoder consumes 6-channel point features and uses five stages with channel widths $[48, 96, 192, 384, 512]$, depths $[3, 3, 3, 12, 3]$, multi-head attention widths $[3, 6, 12, 24, 32]$, patch size 1024, stride-2 downsampling between stages, masked-token support, and spatial ordering based on z-order, transposed z-order, Hilbert, and transposed Hilbert curves. Point features are mapped into the language-model space through an MLP projector.

\paragraph{Sonata.} As a frozen encoder of \textbf{\textcolor{Ground_Color}{\glsplain{LLM-Grounder}}}, we used the \textbf{encoder-only} configuration. The encoder consists of five hierarchical stages with channel dimensions $[48, 96, 192, 384, 512]$, depths $[3, 3, 3, 12, 3]$, with a patch size of 1024 at each stage and stride-2 down-sampling between stages. The model uses pre-normalization, an MLP expansion ratio of 4, learned query/key/value biases, mask tokens, stochastic depth with a drop-path rate of 0.3, and no attention or projection dropout. Flash attention \cite{dao2022flashattention} is enabled for efficient inference, while relative positional encoding, attention upcasting, and softmax upcasting are disabled. Sonata serializes points using multiple space-filling curve orders, including Z-order, transformed Z-order, Hilbert order, and transformed Hilbert order, with order shuffling enabled.
\paragraph{Training Protocol.}
Since ScanRefer, Nr3D, and Sr3D are all built upon the ScanNet dataset, we adopt a two-stage training strategy.

In Stage 1, the model is first pre-trained on the SpatialLM synthetic dataset, then further trained for 3 epochs on the same dataset with class labels mapped to the ScanNet taxonomy, and finally fine-tuned on ScanNet for 30 epochs. All steps in this stage are conducted on the object detection task by~\cite{SpatialLM}.

In Stage 2, the model is fine-tuned on the \glsplain{3DVG} task for 10 epochs with a batch size of 4 on each dataset (ScanRefer, Nr3D, Sr3D) separately. We use the AdamW optimizer with a learning rate of 1e-4, a cosine scheduler, and a warm-up ratio of 0.03; all other hyperparameters remain unchanged. All components (encoder, projector, and LLM) are jointly trainable.

\onecolumn
\section{Pipeline Visualization}
\begin{figure*}[h]
    \centering
    \includegraphics[width=0.9\linewidth]{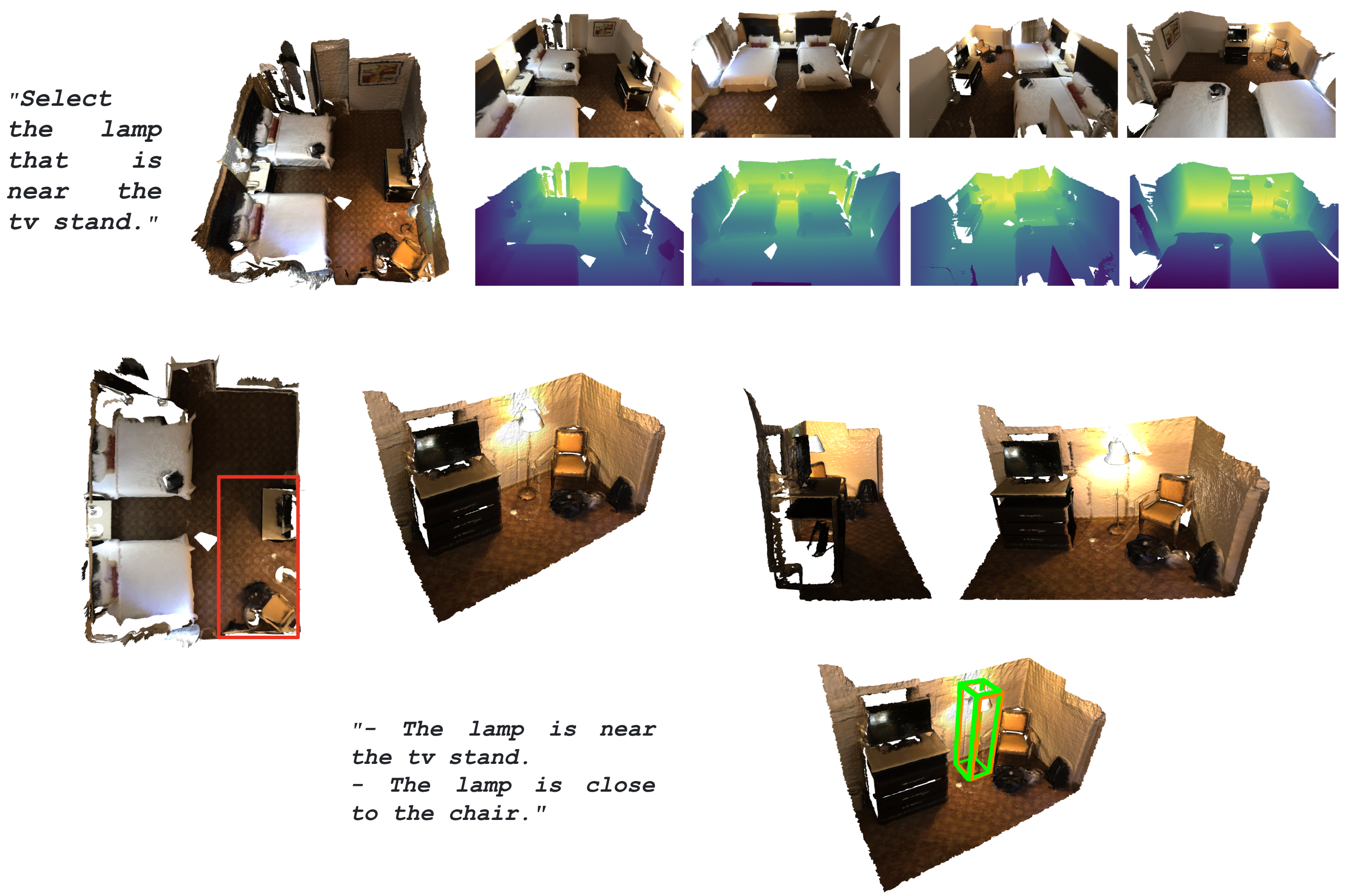}
    \caption{\textbf{Pipeline visualization.}
\\Top row: input description, input scene, and 4 $\times$ oblique views (RGB and depth images) rendered inside \textbf{\textcolor{Prune_Color}{\glsplain{LGSP}}}.
\\Middle row: top view with the 2D bounding box predicted by the frozen VLM, the pruned scene, and two side views.
\\Bottom row: reformulated description generated by \textbf{\textcolor{DescriptReform_Color}{\glsplain{MCDR}}} and the final localization result by \textbf{\textcolor{Ground_Color}{\glsplain{LLM-Grounder}}} 
\\(\textcolor{red}{red}: prediction; \textcolor{green}{green}: ground truth).
    }
    \label{fig:pipeline_visualization}
\end{figure*}

In this section, we provide visualizations of each step in our \PruneGround \,pipeline in Fig.\ref{fig:pipeline_visualization}: (i) \textbf{\textcolor{Prune_Color}{\glsplain{LGSP}}}, (ii) \textbf{\textcolor{DescriptReform_Color}{\glsplain{MCDR}}}, (iii) \textbf{\textcolor{Ground_Color}{\glsplain{LLM-Grounder}}}.

\onecolumn
\section{Failure Case Analysis}
\label{sec:failure_case_analysis}
\begin{figure*}[h]
    \centering
    \includegraphics[width=0.9\linewidth]{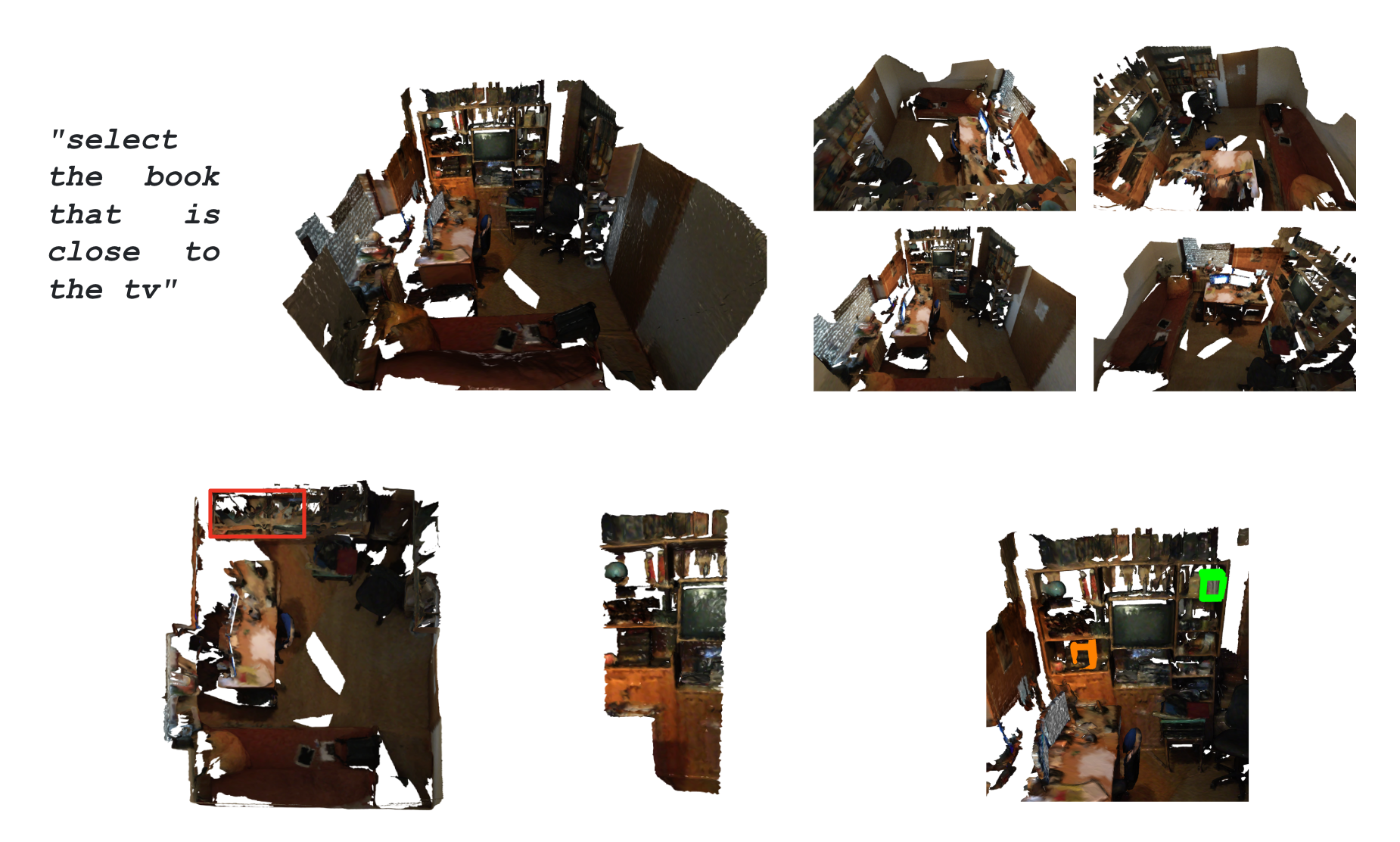}
    \caption{\textbf{Failure case analysis}. \\
    Top row: input description, input scene, and 4 $\times$ oblique views (RGB) rendered inside \textbf{\textcolor{Prune_Color}{\glsplain{LGSP}}}. \\
    Bottom row: top view with the 2D bounding box predicted inside \textbf{\textcolor{DescriptReform_Color}{\glsplain{MCDR}}}, the pruned scene by \textbf{\textcolor{Prune_Color}{\glsplain{LGSP}}}, and 3D bounding box predicted by \textbf{\textcolor{Ground_Color}{\glsplain{LLM-Grounder}}} after mapping to original scene.
\\(\textcolor{red}{red}: prediction; \textcolor{green}{green}: ground truth).
}
    \label{fig:failure_case_analysis}
\end{figure*}

In this section, we discuss the remaining limitations of our proposed framework.
\paragraph{Rendering failures leads to pruning failures.}
First, rendering oblique views of complex indoor scenes remains challenging, particularly under low lighting, with incomplete point clouds, or in multi-room layouts. In such environments, objects are frequently occluded by walls or surrounding furniture, resulting in low-quality renderings. As illustrated in Fig.~\ref{fig:failure_case_analysis}(a), these degraded images can adversely affect the performance of the VLM in identifying language-relevant regions, then lead to final incorrect localization.

\end{document}